%% file: CIKM2025.tex
\documentclass[sigconf]{acmart}
\AtBeginDocument{%
  }

\copyrightyear{2025}
\acmYear{2025}
\setcopyright{cc}
\setcctype{by}
\acmConference[CIKM '25]{Proceedings of the 34th ACM International Conference on Information and Knowledge Management}{November 10--14, 2025}{Seoul, Republic of Korea.}
\acmBooktitle{Proceedings of the 34th ACM International Conference on Information and Knowledge Management (CIKM '25), November 10--14, 2025, Seoul, Republic of Korea}
\acmDOI{10.1145/3746252.3760807}
\acmISBN{979-8-4007-2040-6/2025/11}

\usepackage{microtype}

\usepackage{graphicx}

\usepackage{pgfplotstable}
\usepackage{xfp}
\usepackage{fp}
\usepackage{pgf}
\usepackage{pgfplots}
\pgfplotsset{compat=1.18}
\usepackage{cellspace}
\usepackage{colortbl} 
\usepackage[dvipsnames]{xcolor}
\usepackage{tikz} 
\usepackage{highlight}
\usepackage{url}
\usepackage{subcaption}
\usepackage{amsmath}
\usepackage{tabularx}
\newcolumntype{Y}{>{\raggedright\arraybackslash}X}
\usepackage[export]{adjustbox}
\usepackage{tablefootnote}
\usepackage{CJKutf8}

\input{colors}

\begin{document}

\title{When Language Shapes Thought: Cross-Lingual Transfer of Factual Knowledge in Question Answering}

\author{Eojin Kang}
\email{eojinkang@hufs.ac.kr}
\orcid{0009-0001-5515-5494}
\affiliation{%
  \institution{Hankuk University of Foreign Studies}
  \city{Seoul}
  \country{Republic of Korea}
}

\author{Juae Kim}
\authornote{Corresponding author.}
\email{juaekim@hufs.ac.kr}
\orcid{0000-0001-7826-5226}
\affiliation{%
  \institution{Hankuk University of Foreign Studies}
  \city{Seoul}
  \country{Republic of Korea}
}


\begin{abstract}
Multilingual large language models (LLMs) offer promising opportunities for cross-lingual information access, yet their use of factual knowledge remains highly sensitive to the input language. Prior work has addressed this through English prompting and evaluation, assuming that English-based reasoning is universally beneficial. In this work, we challenge that assumption by exploring factual knowledge transfer from non-English to English through the lens of Language and Thought Theory. We introduce Language-to-Thought (L2T) prompting, which aligns the model’s internal ``thinking'' language with the source of knowledge. Across three languages and four models, L2T consistently outperforms English-based reasoning, reversing the expected advantage of English prompts. Our code is available at \url{https://github.com/GeomeunByeol/Language2Thought}.
\end{abstract}




\maketitle

\section{Introduction}

Multilingual large language models (LLMs) have proven effective at transferring \textit{linguistic} abilities—understanding and generating language acquired via next-token prediction—across languages \cite{cross-lingual-representation-scale,probing-emergence-cross-lingual}. Building on this, recent studies aim to transfer reasoning abilities from English to lower-resource languages through techniques such as direct translation \cite{MGSM} or prompting (e.g., ``Let's think step by step in English'') \cite{Cross-lingual-Prompting-zero,Not_All_Cross-Lingual-Thought}. However, transferring \textit{factual knowledge}, referring to world knowledge stored in the model’s parameters \cite{cloze}, has not been widely studied. Unlike reasoning tasks—where generating intermediate rationales directly leverages English’s linguistic advantages—factual knowledge cannot benefit similarly. As a result, although factual knowledge warrants study from a new perspective, it has yet to be systematically explored. To address this gap, we introduce key concepts rooted in the Language and Thought theory \cite{Sapir, Whorf, Vygotsky, Dissociating, Nature_language_commu_rather_thought} in §\ref{sec:problem_definition}. We make two contributions:

\begin{figure}[t!]
    \centering
    \includegraphics[width=0.7\columnwidth]{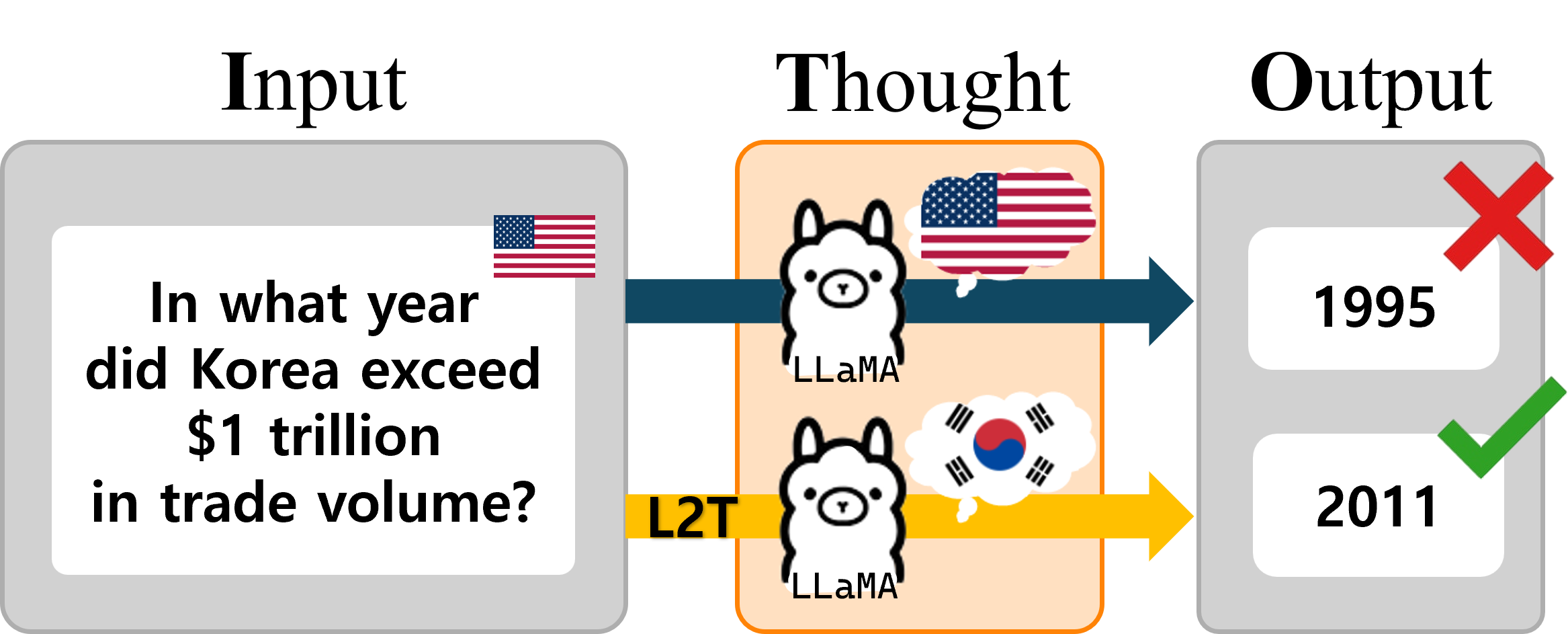}
    \caption{An example illustrating how language influences knowledge transfer. When a question requiring specific factual knowledge about Korea was in English, it led to incorrect responses. Employing our proposed L2T prompting strategy—which shifts the model’s \textit{language of thought}—resulted in correct outcomes by aligning the knowledge required for the task with the internal thinking language.
    }
    \Description{A diagram illustrating the effect of aligning internal thought with the target language in multilingual LLMs. The left panel shows an English input question: "In what year did Korea exceed \$1 trillion in trade volume?" The middle panel labeled Thought has two Llama icons with different flags. The top Llama has an American flag, reflecting English internal parametric process, leading to an incorrect output of 1995. The bottom Llama, influenced by L2T, has a Korean flag, reflecting Korean internal thinking, leading to the correct output 2011. Arrows show the flow of information from Input to Thought to Output.}
    \label{fig:main}
\end{figure}

\textbf{(1) We reverse the traditional direction of knowledge transfer.} While previous studies have focused on transferring English-based knowledge to other languages, instead, we explore knowledge transfer from non-English to English. Our goal is to investigate how data sources influence the limitations of cross-lingual factual knowledge transfer by focusing on the presence and representation of factual data in non-English languages rather than the quantitative predominance of English data.

\textbf{(2) We propose Language-to-Thought (L2T) prompts}, which decouple the parametric process from the input language, based on the hypothesis that LLMs' thought process is closely tied to linguistic form. Whereas standard prompting keeps all stages—input-internal thought-output—in the same language, L2T prompts break this pattern. This subtle flip shifts the model’s thinking language, which in turn alters its ability to utilize factual knowledge.

Through L2T prompting, we analyze performance fluctuations (§\ref{sec:performance}, Figure~\ref{fig:main}) and examine perplexity variations (§\ref{sec:perplexity}). These analyses show how the input language shapes factual knowledge use and highlight the need to understand knowledge characteristics in cross-lingual transfer. We believe these findings offer insights into the language–thought interaction in LLMs and contribute to developing more consistent cross-lingual knowledge transfer strategies.

\input{tables/ITO}

\section{Related Work and Background}\label{sec:problem_definition}
While the relationship between language and thought has played a key role in developing LLMs \cite{Dissociating, CoT}, attempts to explain \textit{factual knowledge} and \textit{multilingualism} within this framework remain scarce. This section addresses these aspects using a Language-Thought perspective. For an extended review of related work, see Appendix~\ref{app:related_work}.
 
\subsection{Language-Thought Binding in LLMs}
LLMs trained with next-token prediction often display a language-binding phenomenon: even when factual information is stored within the parameter space \cite{towards_factual_knowledge}, LLMs generate different answers depending on input language \cite{Sseaeval}. Since language comprehension and the ``thinking'' process (reasoning or knowledge utilization) may function separately \cite{Multilingualism}, we address cross-lingual factual knowledge transfer by decoupling the input language from the internal thinking language. Previous work has primarily addressed this transfer using encoder-based models with smaller, more transparent datasets \cite{Tracing_the_roots, cross-lingual_consistency}. In contrast, by adopting a simple but versatile language-thought binding perspective, we can examine decoder-based LLMs trained on large-scale, opaque corpora that remain difficult to study and underexplored.

\subsection{Language-Thought Conflation Fallacy}
According to \citep{Dissociating}, the language-thought conflation fallacy means that if LLMs are good at language, they must also be good at thinking. This misconception influences cross-lingual transfer research, reinforcing the assumption that since LLMs are most fluent in English, they will also achieve superior performance across all tasks in English. While English does excel in many benchmarks and reasoning tasks \cite{is_translation_all_you_need, bilingual}, overlooking the point that factual knowledge transfer depends more on the availability of relevant information in a given language in training data \cite{Tracing_the_roots} than on linguistic fluency can limit fair and meaningful cross-lingual studies. We investigate factors beyond dataset size that influence how factual knowledge is represented and transferred within LLMs.

\section{Language-to-Thought (L2T) Prompt}\label{sec:method}
As shown in Table~\ref{tab:prompt}, we formalize how the input language influences LLMs' internal thought processes using the notation \texttt{I}:\texttt{{LC}}-\texttt{T}:\texttt{LC}-\texttt{O}:\texttt{LC}, where \texttt{I} represents the Input, \texttt{T} denotes the Thought, and \texttt{O} indicates the Output. \texttt{LC} serves as a placeholder for a specific language code. As a baseline (I:\protect\langbox{gray}{LC1}), we utilize a language-specific dataset, either in its original language or translated into English using the Google Translation API \cite{google_translate}.

In the following settings, we incorporate a ``Think in \{\texttt{LC}\}'' template into the system prompt, corresponding to the \texttt{T}:\texttt{LC} and \texttt{O}:\texttt{LC} components. This setup examines whether specifying a particular language for internal thought influences the dependency between language and factual knowledge. While structurally similar to zero-shot CoT prompting \cite{zeroshot_CoT, reasoning_prompting_survey}, our design serves a distinct purpose: rather than eliciting step-by-step reasoning, we isolate the influence of linguistic framing on factual inference. To this end, we exclude ``step-by-step'' rationale instructions, enabling a direct examination of the relationship between language and factual knowledge.

\paragraph{L2T-Consistent (I:\protect\langbox{gray}{LC1}-T:\protect\langbox{gray}{LC1})}
This setting only instructs the model to think in the same language as the input, serving as a minimal L2T variant for comparison with the baseline. This starting configuration lets us measure how later L2T variants diverge from the model’s original input-dependent behavior.

\paragraph{L2T-Transfer (I:\protect\langbox{gray}{LC1}-T:\protect\langbox{Orange}{LC2})}
This setting introduces an explicit mismatch between the input and the thought language to assess how such a shift impacts parametric factual knowledge. For example, the model may be prompted to ``think in English'' while receiving a non-English input. This configuration allows us to examine how the thought language affects the model's knowledge utilization and whether it impacts the accuracy of responses.

\paragraph{L2T-Align (I:\protect\langbox{gray}{LC1}-T:\protect\langbox{Orange}{LC2}-O:\protect\langbox{gray}{LC1})}
Since L2T-Transfer prompts the model to think in a different language, it may treat this shift as a signal to respond in that language, unintentionally diverging from the input. However, in real-world multilingual applications, maintaining consistency between input and output languages is often crucial \cite{MLLM_survey, improving_ICT_Cross-lingual_alignment}. Therefore, we add an explicit description, ``answer in \{LC1\}'' instruction, guiding the model to separate the thinking and response languages while enforcing input–output alignment. This configuration enables analysis of how linguistic knowledge and factual knowledge interact in multilingual settings.

In summary, L2T is a method for shifting the model’s \textit{internal} thought language. A comparative approach that mimics \textit{external} linguistic and behavioral patterns through persona approach is discussed in Appendix~\ref{app:persona}.

\section{Experimental Setup}
\subsection{Language Selection and Datasets}
We select multiple-choice QA datasets designed for each language — Chinese simplified (ZH) from CMMLU \cite{CMMLU} and SeaEval (SE) \cite{Sseaeval}, Korean (KO) from CLIcK \cite{Click}, and Arabic (AR) from ArabicMMLU \cite{Arabicmmlu}. To ensure relevance and comparability of factual content, we filtered the datasets by \textit{topic}, focusing on fact-based questions rather than linguistic aspects such as grammar or vocabulary, yielding 1,060, 1,345, and 1,466 questions for ZH, KO, and AR, respectively. Our goal is to investigate reverse transfer—from non-English to English—so we chose languages that differ substantially from English in data distribution,\footnote{According to Common Crawl statistics, these languages exhibit varying levels of multilingual web content (ZH: 4.88\%, KO: 0.79\%, AR: 0.67\%) \cite{commoncrawl}, reflecting a spectrum from relatively higher-to-lower-resource conditions.} script, and linguistic structure. While earlier studies suggest that cross-lingual transfer is more effective among alphabetically similar languages (e.g., Latin-script languages) \cite{match_the_script, gpt4}, we focus on the more challenging scenario of transferring factual knowledge across typologically diverse languages.

\subsection{Models and Evaluation}
We analyze four models: Mistral-Small-24B-Instruct 4-bit quantized (Mistral) \cite{mistral}, LLaMA 3.1-70B-Instruct quantized version of GGUF (Llama) \cite{Llama3_1}, GPT-4o-mini (GPT) \cite{openai_gpt4o_mini_2024}, and Gemini 2.0 Flash-Lite (Gemini) \cite{gemini_lite}. These models were selected based on their multilingual capabilities and model sizes. For more details on the model selection criteria and analysis, refer to Appendix~\ref{app:model_analysis}. For extended results, including models not covered in the main text, refer to Appendix~\ref{app:all_results}. 

We adopt a zero-shot setting to evaluate LLMs’ inherent capabilities in reasoning, preventing factual knowledge from being influenced by in-context examples. For all models, we set the maximum new token limit to 1024 and used each model’s default temperature and top-p values. We report average accuracy over three runs and use the LangDetect to detect response language \cite{langdetect}.

\input{tables/baseline_final}
\input{tables/I_T_acc_final}

\section{Impact of L2T on Knowledge-Focused QA}\label{sec:performance}
In this section, we investigate the relationship between input language (I), internal thought (T), and factual knowledge by evaluating performance on knowledge-based QA tasks. 

\subsection{Multilingual LLMs Are Locally Monolingual}\label{sec:consistent}
The baseline results in Table~\ref{tab:baseline} show that translating inputs into English does not guarantee better performance. GPT and Gemini consistently outperform their English-translated versions of non-English inputs, and even Llama and Mistral—despite limited multilingual tuning—tend to achieve higher accuracy in the original language. Specifically, Llama, which is not instruction-tuned for these languages, surpasses its English translations in ZH and KO by 4.87\% and 2.75\%, respectively. These findings diverge from previous studies that suggest translating into English improves performance \cite{Do_multilingual_Language_Models_Think_better_eng, bilingual}. Our results raise the possibility that English’s superior performance in past research is not solely due to its larger data volume or linguistic capabilities, but rather because general knowledge has been more extensively documented in English.

To understand LLMs' dependency on the input language, we further examine the model’s process in the L2T-Consistent setting. If the performance of the \texttt{I:EN-T:EN} configuration improves, it would indicate an effect similar to zero-shot CoT. However, as shown in Table~\ref{tab:Consistent_Acc}, performance remains nearly unchanged compared to the baseline (Table~\ref{tab:baseline}), with only minimal shifts in accuracy. This suggests that the internal reasoning process remains closely tied to the input language, motivating the transition to more explicit strategies that guide internal thought.

\input{images/perplexity}
\input{tables/I_T_IN_final}
\input{tables/equal}

\subsection{Can LLMs Think in a Different Language?}\label{sec:transfer}
Here, we discuss the results of attempts to alter the model's internal thought process using L2T-Transfer. Llama consistently exhibited a notable performance drop, likely due to its relatively weaker ability to follow multilingual instructions. Therefore, we primarily focus on the results from the other models, which demonstrate stronger multilingual generation capabilities.

As shown in Table~\ref{tab:Transfer_Acc}, across all \texttt{I:Orig-T:EN} configurations, performance declined (red cells), suggesting that the improvements reported in previous cross-lingual prompting studies \cite{Cross-lingual-Prompting-zero, Not_All_Cross-Lingual-Thought} may not apply universally. These findings are in accord with previous results that English-based prompting does not always improve performance. In contrast, the \texttt{I:EN-T:Orig} settings yielded modest gains (blue cells). This indicates that misalignment between the task and internal thought often disrupts parametric knowledge access, whereas aligning task requirements with internal thought processes can enhance knowledge activation.

In addition to accuracy, we examine how L2T-Transfer influences the response language (Table~\ref{tab:I_T_IN}). In the \texttt{I:EN-T:Orig} setting, we observe a noticeable increase in outputs in the guided language, suggesting that working in non-English often leads to unintended language shifts. Conversely, in the \texttt{I:Orig-T:EN} setting, the response language largely follows the input language. The model tends to maintain the intended output language more consistently, indicating a more stable reasoning process and fewer unintended language shifts. These findings highlight the importance of aligning the input language with the model's external reasoning processes to enhance language control and factual consistency, which we investigate further in the next section.

\subsection{Optimizing Language Alignment for Effective Knowledge Utilization}\label{sec:align}
Table~\ref{tab:equal} presents the results of the L2T-Align configuration, where the output language predominantly aligns with the input language (Table~\ref{tab:Transfer_In}). This suggests that the external reasoning process remains comparable to the baseline. However, performance variations arise due to differences in how internal knowledge is activated.

Performance Gains in Alignment Setting (\protect\bodybox{gray}{EN}-\protect\langbox{Orange}{Orig}-\protect\bodybox{gray}{EN}).
Across all models and languages, the model consistently gains access to factual knowledge primarily encoded in the original language, facilitating more effective knowledge transfer (blue cells). For example, Gemini shows an average performance improvement of 1.99\%. This improvement highlights how L2T-Align mitigates disruptions to the external reasoning process seen in L2T-Transfer by aligning linguistic context more closely. This setting also represents a realistic scenario in LLM usage, where the input and output languages remain the same. For an example of the model response, see Appendix \ref{app:cherry}. 

A notable exception occurs with Gemini under \texttt{I:EN-T:ZH-O:EN}, where 27.52\% of the response appears in intermediate descriptions in ZH, while only the final answer appears in EN. This may reflect Gemini’s stronger adherence to internal language cues, as emphasized in official system instruction guidelines \cite{gemini_system_instruction}, or a ZH-specific tendency resulting from increased rationale-style supervision (e.g., Deepseek-R1 \cite{deepseek}).

Performance Decline in Misalignment Setting (\protect\langbox{gray}{Orig}-\protect\bodybox{Orange}{EN}-\protect\langbox{gray}{Orig}).
This setting, like \texttt{I}:\texttt{EN}-\texttt{T}:\texttt{Orig}-\texttt{O}:\texttt{EN}, also includes a dual-language specification for internal thought and output. However, instructing the model to adopt EN for its internal thinking causes a misalignment, impairing its ability to effectively utilize factual knowledge, thereby leading to a performance decline (e.g., 1.71\% on average in Gemini). Overall, these results indicate that effectively aligning the input language, internal thought, and output language enhances the model's ability to elicit latent knowledge. This emphasizes the importance of prompt-level language coordination in multilingual LLMs. While the focus here was on non-English to English knowledge transfer, the potential for non-English to non-English transfer is discussed in Appendix~\ref{app:NonEN_to_NonEN}.

\section{Measuring Language-Thought Alignment}\label{sec:perplexity}
To further investigate how language–thought alignment affects model behavior, we analyze perplexity (PPL) as a metric for familiarity and certainty. Following prior work linking PPL to pretraining exposure and data coverage \cite{Detecting_Pretraining, Demystifying_Prompts_Perplexity}, we use PPL to validate the parametric effect of L2T prompting. To reduce noise and focus on stable behavior, we consider questions that are answered either correctly or incorrectly across all three runs and report the average PPL. Definitions and computation details are provided in Appendix~\ref{app:perplexity}.

As shown in Figure~\ref{fig:perplexity}, PPL is consistently lower when the input is in the original language (\textit{I:Orig}) compared to English (\textit{I:EN}), regardless of L2T configuration. This trend contradicts the common assumption that higher-resource languages produce more confident predictions than lower-resource ones \cite{multilingual_perplexity} (further explanation in Appendix \ref{app:mmlu_perplexity}). This irregularity suggests that the alignment of language, thought, and task plays a crucial role in the model’s familiarity with the input. Overall, these PPL results reinforce our central claim: effective knowledge utilization is not merely a function of language size or training frequency, but depends critically on how language, thought, and task requirements are aligned during inference.

We also observe consistent variations across L2T settings. When the input is in the original language, L2T-Consistent yields lower PPL than L2T-Transfer or L2T-Align, indicating smoother internal processing. Conversely, when the input is in English, L2T-Transfer and L2T-Align often reduce PPL relative to L2T-Consistent. These trends imply that L2T prompts influence not only final accuracy but also the model’s internal computational pathways.

\section{Conclusion}
This paper explores how LLMs’ ability to elicit knowledge is constrained by the input language through Language and Thought Theory and proposes L2T prompting. Experimental results reveal that effective cross-lingual knowledge transfer requires jointly aligning the input and output languages as well as the internal language of thought. For future work, we aim to further expand training-time strategies to internalize multilingual reasoning and support the integration of knowledge encoded in diverse languages.


\section*{Limitations}

This study focuses on factual knowledge transfer, and we have made efforts to extract and analyze data primarily targeting factual knowledge. However, certain non-factual factors can be inevitably involved, even in factual tasks. For example, some questions inherently require basic linguistic understanding (e.g., ``Which of the following is not a traditional Chinese outfit?'') or necessitate reasoning to arrive at the correct answer \cite{knowledge_reasoning}. This complexity has led previous studies to focus on cloze-style tasks for factual knowledge \cite{cloze}; however, such formats pose challenges when applied to recent large, decoder-based multilingual language models \cite{factual_knowledge_cloze_distractor}.
For this reason, we adopted QA-based datasets, aligning with emerging research trends that approach factual knowledge transfer through question-answering formats \cite{towards_factual_knowledge}. Our study focuses on factual knowledge that is both culturally grounded and language-specific, which led us to limit the task format to multiple-choice questions (MCQs). Given the current lack of multilingual open-ended QA resources, expanding this research to include more diverse task formats and datasets would be a valuable direction for future work.

While we aimed to cover a diverse range of languages, it was not feasible to experiment with all languages. However, as our findings suggest, language-specific characteristics may influence model behavior, and these differences could stem from the way LLMs encode and process different languages. Notably, despite all models being multilingual by design, we observed meaningful variations in performance and response patterns across different languages. Further comparative analyses of language-specific properties in LLMs are needed to better understand how these variations impact cross-lingual factual knowledge transfer.

\begin{acks}
This work was supported by the National Research Foundation of Korea (NRF) grant funded by the Korea government (MSIT) (No. RS-2025-24523068).
\end{acks}

\section*{GenAI Usage Disclosure}
We used generative AI tools solely for the purpose of editing and improving spelling, grammar, punctuation, and clarity. All ideas and content in this manuscript remain entirely the work of the authors.

\input{reference.bbl}


\appendix
\newpage

\begin{table}[t!]
  \centering
  \begin{tabular}{l}
    \hline
    \textbf{Instructions} \\
    \hline
    Answer the given multiple choice question and show your work. \\ The answer can only be an option like A, B, C, or D. \\ You need to output the answer in your final sentence like \\ ``Therefore, the answer is ...''. \\
    \hline
  \end{tabular}
  \caption{English instructions.}
  \label{tab:user_prompt_en}
\end{table}

\begin{table}[h!]
  \centering
  \begin{tabular}{l}
    \hline
    \textbf{Instructions} \\
    \hline
    \begin{CJK*}{UTF8}{gbsn}回答给定的多项选择题并展示您的解题过程。\end{CJK*} \\ \begin{CJK*}{UTF8}{gbsn}答案只能是一个选项，如A、B、C或D \end{CJK*} \\ \begin{CJK*}{UTF8}{gbsn} 您需要在最后一句中输出答案 如“因此，\end{CJK*} \\ \begin{CJK*}{UTF8}{gbsn} 答案是...”。\end{CJK*} \\
    \hline
  \end{tabular}
  \caption{Simplified Chinese instructions.}
  \label{tab:user_prompt_zh}
\end{table}

\begin{table}[h!]
  \centering
  \begin{tabular}{l}
    \hline
    \textbf{Instructions} \\
    \hline
    \begin{CJK}{UTF8}{mj}주어진 객관식 문제에 답하고, 풀이 과정을 보이시오. \end{CJK} \\ 
    \begin{CJK}{UTF8}{mj} 정답은 A, B, C, D 중 하나여야 한다. \end{CJK} \\
    \begin{CJK}{UTF8}{mj} 마지막 문장에서 ``따라서 답은 ...''처럼 답을 출력하시오.\end{CJK} \\
    \hline
  \end{tabular}
  \caption{Korean instructions.}
  \label{tab:user_prompt_ko}
\end{table}

\begin{table}[h!]
  \centering
  \begin{tabular}{l}
    \hline
    \textbf{Instructions} \\
    \hline
    \includegraphics[width=\linewidth]{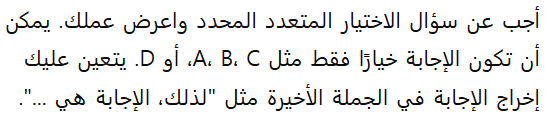} \\
    \hline
  \end{tabular}
  \caption{Arabic instructions.}
  \label{tab:user_prompt_ar}
\end{table}

\begin{table*}[t!]
  \centering
  \renewcommand{\arraystretch}{1.2}
  \begin{tabular}{c|c|c|c}
    \hline
     & \textbf{L2T-Consistent} & \textbf{L2T-Transfer} & \textbf{L2T-Align} \\
    \hline
    \textbf{EN} & Think in English. & Think in \{Orig\}. & Think in \{Orig\} and answer in English. \\
    \hline
    \textbf{ZH} &
      \begin{CJK*}{UTF8}{gbsn}用中文思考。\end{CJK*} &
      \begin{CJK*}{UTF8}{gbsn}用英语思考。\end{CJK*} &
      \begin{CJK*}{UTF8}{gbsn}用英语思考，用中文回答。\end{CJK*} \\
    \hline
    \textbf{KO} &
      \begin{CJK}{UTF8}{mj}한국어로 생각하시오.\end{CJK} &
      \begin{CJK}{UTF8}{mj}영어로 생각하시오.\end{CJK} &
      \begin{CJK}{UTF8}{mj}영어로 생각하고 한국어로 답하시오.\end{CJK} \\
    \hline
    \textbf{AR} &
      \includegraphics[width=0.13\linewidth]{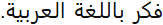} &
      \includegraphics[width=0.15\linewidth]{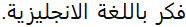} &
      \includegraphics[width=0.31\linewidth]{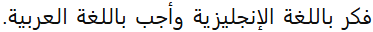} \\
    \hline
  \end{tabular}
  \caption{L2T prompts for each input language.}
  \label{tab:system_prompt}
\end{table*}

\section{Additional Literature Review}\label{app:related_work}
\subsection{Factual Knowledge}
In this work, we adopt the perspective that language models store factual knowledge and perform reasoning based on it, treating factual knowledge and reasoning as distinct components \cite{reasoning_survey}. Since factual knowledge is broadly applicable across tasks, it is important to understand how much and how of it is naturally acquired during the pretraining stage \cite{how_factual_knowledge}. Common evaluation setups include cloze tasks \cite{cloze}, question answering datasets \cite{mcq_exam, towards_factual_knowledge}, and metrics such as P@1, log probability, or perplexity \cite{factual_ppl}. However, most have been applied to encoder-based pretrained language models with clear access to training data, whereas research on decoder-based LLMs—typically trained on massive and unreleased corpora—remains limited and challenging.

\subsection{Multilingual Large Language Model and Cross-Lingual Transfer}
With the rise of multilingual language models, low-resource languages increasingly benefit from the rich datasets available in English \cite{MLLM_survey, cross_lingual_survey}. This has led to an active body of research on cross-lingual transfer, including methods such as translating low-resource inputs into English \cite{Do_multilingual_Language_Models_Think_better_eng, is_translation_all_you_need}, and cross-lingual prompting or tuning strategies \cite{Cross-lingual-Prompting-zero, Not_All_Cross-Lingual-Thought, crosslingual_tuning}. However, these efforts have mainly focused on transferring reasoning or linguistic knowledge, which are more directly tied to language, while transferability in language-agnostic factual knowledge remains underexplored.

\subsection{Factual Knowledge in LLMs}
Although leveraging knowledge bases expressed in multiple languages is critical \cite{zebra}, LLMs still exhibit performance inconsistencies depending on the input language \cite{Sseaeval, cross-lingual_consistency}. Prior work on knowledge transfer in multilingual settings has largely focused on encoder-based models with smaller and more transparent datasets \cite{Tracing_the_roots, cross-lingual_consistency}. In contrast, factual knowledge transfer in decoder-based LLMs—trained on massive, opaque corpora—has been difficult to study and remains underexplored. Our work shifts the focus to factual knowledge transfer from non-English to English, and investigates factors beyond dataset size that influence how factual knowledge is represented and transferred within LLMs.

\input{tables/prompt}
\section{Prompt Examples}\label{app:prompt}

Table~\ref{tab:user_prompt_en}, Table~\ref{tab:user_prompt_zh}, Table~\ref{tab:user_prompt_ko}, and Table~\ref{tab:user_prompt_ar} present the instruction prompts added before each question. Table~\ref{tab:system_prompt} shows our L2T prompts corresponding to each input language.

\section{Model Analysis}\label{app:model_analysis}
\input{tables/model_analysis}

In this work, we selected models for evaluation based on two main criteria. First, we included strong closed-source models known for their high multilingual capabilities, such as GPT-4o-mini, Gemini-2.0-Lite, and Mistral-Small-24B. Second, we considered open-source models like Llama, which, despite lacking explicit instruction tuning for multilingual tasks, were pretrained on multilingual corpora. Additionally, we aimed to investigate two important dimensions: the effect of improved multilinguality by comparing Llama3.0 and Llama3.1, and the impact of model scale by comparing 8B and 70B parameter versions.

Table \ref{tab:model_analysis} presents the performance of each model on MMLU (English) \cite{mmlu, mmlu_pro} and MMMLU (multilingual) \footnote{\url{https://huggingface.co/datasets/openai/MMMLU}}, two representative benchmarks for evaluating factual knowledge and reasoning capabilities.

\section{Dataset Details}\label{app:dataset}

In line with our goal of examining factual knowledge transfer and its dependence on language-specific knowledge availability, we focus on the reverse transfer direction, investigating the transfer of factual knowledge from non-English languages into English. This approach requires careful selection of the dataset.

To focus on factual knowledge rather than linguistic aspects such as grammar or vocabulary, we filtered the datasets by excluding linguistic knowledge and prioritizing QA pairs centered on factual information. This filtering is crucial because the process of translating linguistic questions can alter their meaning, but our goal is not to measure performance degradation caused by translation. 

The filtered topics are as follows:
\begin{itemize}
    \item ZH: Food Culture, Foreign Policy, History, Literature, Traditional Chinese Medicine, etc.
    \item KO: Economy, Geography, History, Law, Politics, Popular Culture, Society, Tradition, etc.
    \item AR: Islamic Studies, etc.
\end{itemize}

\section{Supplementary Model Experiment Results}\label{app:all_results}
In this section, we present the results of additional model experiments beyond those discussed in the main text, which primarily focused on Llama3.1-70B-Instruct and GPT-4o-mini. Here, we extend our analysis to include results from Llama3-8B-Instruct, Llama3.1-8B-Instruct, Llama3-70B-Instruct, and GPT-3.5-turbo.

\subsection{Baseline Results}
Table~\ref{tab:baseline_appendix} presents the baseline results. For ZH and KO, performance consistently remained better in the original language than in its English translation across almost all models. A notable finding is that Llama3 models, known for their limited multilingual abilities, predominantly generated responses in English even when the input language was ZH or KO. Despite this tendency, the models still performed better when the input was in the original language. This unexpected pattern may suggest that while the model's ability to retrieve factual knowledge differs depending on the input language, the knowledge itself remains independent of the model's linguistic generation abilities. In other words, factual knowledge recall and linguistic generation may be governed by distinct mechanisms.

\subsection{L2T-Consistent Results}\label{app:THC}
Table~\ref{tab:mix_same_appendix} shows the results for the L2T-Consistent setting. Unlike the baseline, where Llama3 models predominantly responded in English even when the input was in ZH or KO, the response language more frequently aligned with the input language under this configuration. We also observed that models with better multilingual capabilities and stronger instruction-following performance were more adept at adhering to the L2T prompt instructions, resulting in smaller performance gaps compared to the baseline.

\subsection{L2T-Transfer Results}
Table~\ref{tab:mix_different_appendix} presents the results for the L2T-Transfer setting. In this configuration, models with weaker multilinguality or less reliable instruction-following abilities often experienced difficulties in engaging in language-specific thought retrieval. Conversely, more capable models exhibited signs of successful cross-lingual knowledge transfer.

\input{tables/baseline_appendix}

\input{tables/mix_same_appendix}

\input{tables/mix_different_appendix}

\input{tables/equal_appendix}

\subsection{L2T-Align Results}
Table~\ref{tab:equal_appendix} displays the results for the L2T-Align setting. As observed, models with stronger multilinguality and better instruction-following abilities exhibited better task performance when the input language, internal thought language, and task requirements were aligned. In the \textit{I:Orig-T:EN-O:Orig} configuration, which aimed to suppress knowledge retrieval by misaligning the internal thought language with the task, the average performance decreased by 0.51\% for Llama3-70B, 0.42\% for Llama3.1-70B, and 0.63\% for GPT-4o-mini. In contrast, in the \textit{I:EN-T:Orig-O:EN} configuration, which aimed to facilitate knowledge transfer by aligning internal thought with the task language, performance increased by 0.34\% for Llama3-70B, 1.17\% for Llama3.1-70B, and 0.74\% for GPT-4o-mini.

\section{Comparative Evaluation: Persona}\label{app:persona}
\input{tables/persona_prompt}

\input{tables/persona_consistent}
\input{tables/persona_transfer}

In this work, we proposed L2T as a method for influencing the internal thought process. In this section, we present a comparative experiment using a persona-based approach as an alternative to L2T. The persona-based approach is designed to simulate human-like characteristics by mimicking linguistic and behavioral patterns \cite{wang2024incharacter}. While the persona-based method can be considered part of the broader L2T framework because it alters the thinking language, it includes more explicit linguistic instructions, making it more about changing the external language, whereas L2T focuses on manipulating the internal language.

We adopted the persona prompt structure from \cite{persona_bias} as shown in Table~\ref{persona_instruction_en}. The placeholder \{persona\} is filled with descriptions like ``an English speaker'' or ``an Arabic speaker''. 

When the input language matches the assigned persona, the setting corresponds to the L2T-Consistent condition, which we refer to as Persona-Consistent (e.g., \textit{I:EN-T:an English speaker}). Conversely, when the persona differs from the input language, the setting corresponds to the L2T-Transfer condition, which we call Persona-Transfer (e.g., \textit{I:EN-T:an Arabic Speaker}). The translations of the English prompts into non-English prompts can be found in Table~\ref{persona_instruction_zh}, Table~\ref{persona_instruction_ko}, and Table~\ref{persona_instruction_ar}. The experiment involved three different persona assignment methods, each run three times, resulting in a total of nine runs for the final results.

Table~\ref{tab:persona_consistent} shows the results for the Persona-Consistent configuration. We observed a decrease in the number of responses generated in English when the input was in ZH or KO, compared to the results reported in Section~\ref{app:THC}
. This suggests that the persona-based prompts effectively mimic the linguistic and behavioral patterns of the assigned persona. However, the performance differences were minimal (GPT-4o-mini Baseline: ZH: 57.67, KO: 69.62, AR: 79.85 / Persona-Consistent: ZH: 57.71, KO: 69.16, AR: 79.4), resembling the findings in Section~\ref{sec:consistent}, where we analyzed the consistency between the input language and the internal thought process.

Table~\ref{tab:persona_transfer} shows the results for the Persona-Transfer configuration. We found that performance often improved when using the \textit{I:EN-T:Orig} setting, while performance occasionally dropped in the \textit{I:Orig-T:EN} setting. This result further supports the importance of correctly assigning the internal thought language for effective factual knowledge retrieval. However, unlike L2T prompts, persona-based prompts contain more explicit linguistic instructions. As a result, the phenomenon we observed in Section~\ref{sec:transfer}, where the model interprets the instruction as a directive to switch the output language, became more pronounced. This issue was particularly noticeable with Llama3.1, which frequently produced phonetic transliterations of the original language into the Latin alphabet instead of responding correctly. These results suggest that persona-based prompting may focus more on linguistic generation than on the internal cognitive process required for knowledge retrieval. In practical, real-world scenarios where input and output language consistency is critical, the Think-Align approach appears to be a more reliable solution.

\section{Non-EN to Non-EN Knowledge Transfer}\label{app:NonEN_to_NonEN}
\input{images/NonENtoNonEN}

In Section~\ref{sec:performance}, we demonstrated that the type of knowledge and task alignment play a critical role in factual retrieval, and that English-based cross-lingual transfer does not always guarantee improved performance. Extending this line of analysis, we further explore cross-lingual transfer in a non-English-to-non-English setting.

Table~\ref{fig:NonEN_to_NonEN} presents the results of the L2T-Align configuration (I:\langbox{gray}{LC1}-T:\langbox{orange}{Orig}-O:\langbox{gray}{LC1}), where aligning non-English input and output languages with the appropriate internal thought language leads to noticeable performance gains. These findings reaffirm that language ability is not solely driven by data scale but also by the interaction between thought and knowledge, pointing toward the potential for more democratic LLMs and broader cross-lingual transfer beyond English.

\section{PPL Definition}\label{app:perplexity}
The PPL calculation used in Section~\ref{sec:perplexity} is defined as follows:
\begin{equation}
\text{PPL} = \exp\left( -\frac{1}{m-k+1} \sum_{i=k}^m \log P(x_i \mid x_{<i}) \right)
\end{equation}

\begin{itemize}
    \item \( k \): The index of the starting token in the user prompt (including instructions, a question, and options).
    \item \( m \): The index of the ending token in the user prompt.
    \item \( X_{\text{user}} = (x_k, x_{k+1}, \ldots, x_m) \): The sequence of tokens in the user prompt.
\end{itemize}

While our L2T system prompts influence the model’s PPL indirectly—by affecting the predicted token distributions—, they are not included in the direct computation of PPL. In other words, PPL is calculated only over the tokens in the user prompt, excluding the L2T system prompt itself. This is because our goal is not to measure the PPL of the prompt itself but rather to analyze how the model responds to factual questions under the influence of the L2T prompts.

\section{Comparative Evaluation: MMLU}\label{app:mmlu_perplexity}
\begin{table}[t]
    \centering
\begin{tabular}{l|c|c}
\hline
\textbf{Input}                       & \textbf{EN} & \textbf{KO} \\
\hline
\textbf{Baseline}      & 9.96        & 11.52       \\ 
\hline
\textbf{L2T-Align} & 10.63       & 10.23      \\ 
\hline
\end{tabular}
    \caption{PPL results on the MMLU dataset.}\label{tab:mmlu}
\end{table}

In Section~\ref{sec:perplexity}, we observed an unusual trend where the PPL for English inputs was consistently higher than for non-English inputs when dealing with specialized factual knowledge. This finding contradicts the common assumption that higher-resource languages generally yield more confident predictions than lower-resource ones. To further investigate this phenomenon, we conduct an additional analysis focusing on world knowledge, which is more universally shared across languages.

We randomly sampled 1,000 questions from the MMLU benchmark \cite{mmlu}. In our previous experiments, we used datasets originally written in non-English languages and subsequently translated into English using machine translation. Such a translation process could potentially influence the PPL results. To address this concern and ensure consistency in Orig to EN translation direction, we utilized the Korean portion of the MMMLU dataset, \textit{human-translated} from English into multiple languages. and then translated it back into English using the same machine translation process applied in earlier experiments.

Table~\ref{tab:mmlu} shows the PPL results for both Baseline and L2T-Align configurations, comparing English and Korean data. As expected, the results for world knowledge align more closely with conventional assumptions regarding language resource availability. In Baseline Configuration, English PPL is lower than Korean PPL, indicating that the previously observed perplexity trends are not merely artifacts of machine translation. This result supports our previous findings that internal predictions become more stable when knowledge is more extensively expressed in the target language during training.

In the L2T-Align configuration, the PPL in the \textit{I:EN-T:Orig-O:EN} setting increased by 0.67 compared to the baseline. In contrast, in the \textit{I:Orig-T:EN-O:Orig} setting, perplexity decreased by 1.29 relative to the baseline. These results suggest that internal thought language can affect knowledge retrieval, consistent with our earlier findings. Overall, the experimental results suggest that internal thought language may play a role in knowledge retrieval and cross-lingual transfer. Through this additional experiment, we aimed to further emphasize that our intention was not to demonstrate a unidirectional transfer of knowledge, but rather to investigate the underlying factors that influence cross-lingual knowledge transfer more broadly.

\section{Knowledge Transfer Enhancement with L2T}\label{sec:tuning}

While the body of this paper demonstrated the role of L2T prompting in guiding cross-lingual thought, we further ask: can language-specific thought patterns be internalized through training? We conduct a preliminary experiment to explore whether models can internalize language-specific thought patterns purely through exposure to additional data, without explicit task-specific fine-tuning. We extend Llama3.1-70B-Instruct's training using QLoRA \cite{QLoRA} on a Korean-specific dataset (100K factual statements related to Korean culture) \cite{AIhub}. These statements are not directly related to the evaluation datasets but serve purely as additional exposure during continued pretraining. After training, we evaluate the model using the same factual knowledge benchmarks used in previous sections. We experiment with four different configurations: (1) Original Data Only – known as zero-shot transfer \cite{zero_transfer}, (2) Translated English Data – effective but resource-intensive, (3) Original Data + L2T (KO) Prefix – appending ``Let’s think in Korean'' to guide thought alignment, and (4) Original Data + L2T (EN) Prefix – counterpart of (3).

Table~\ref{tab:tuning} shows the results for each setting. In the (1) setting, no significant improvement was observed in either KO or EN performance. This suggests that the model might already possess the relevant factual knowledge in KO, and zero-shot transfer from KO to EN remained ineffective. In contrast, the (2) setting yielded a 1.83\% improvement in EN but a 1.27\% drop in KO performance. This indicates that while translation-based training effectively adds knowledge in the target language, it may also negatively impact performance in other languages. A notable finding emerged from the (3) setting, which achieved a 1.64\% improvement—a result comparable to translation-based training and earlier prompting experiments. This indicates that guiding the model’s internal thought process during training can partially overcome LLMs' limitations, which stem from their tendency to remain tightly bound to the input language and struggle with cross-lingual knowledge transfer. Furthermore, these results encourage further consideration of ways to facilitate knowledge transfer during training using only the original data, rather than relying on resource-intensive and potentially harmful translation-based methods.
\input{tables/tuning_result}
\section{Representative Response Example}\label{app:cherry}

\input{tables/cherry}
Figure~\ref{fig:main} and Table~\ref{tab:cherry} present a sample QA from the CLIcK benchmark \cite{Click}, which involves Korean-specific factual knowledge. The Llama3.1-70B-Instruct consistently failed to answer this question under the Baseline setting (\textit{I:EN}) across three runs but answered correctly in all runs with the L2T-Align setting (\textit{I:EN-T:KO-O:EN}).

The question was: ``Why are there so many people traveling between Seoul and Gyeonggi-do every day?'' 

Under the Baseline setting, the model responded with: ``Because they are close.'' This response shows a guess based on geographic proximity, resulting in a plausible but inaccurate answer.

In contrast, under the L2T-Align setting, the model responded with: ``For work, study, etc.'' This demonstrates that the model provided a more accurate answer based on factual knowledge specific to Korea, as Seoul is a central hub for various activities.

\end{document}

%% file: colors.tex
\newcommand{\relang}[1]{\gradientcellreverse{#1}{1}{100}{white}{PineGreen}{70}}

\newcommand{\kooko}[1]{\gradientcelld{#1}{67.62}{69.62}{71.62}{red}{white}{blue}{50}}
\newcommand{\kooen}[1]{\gradientcelld{#1}{60.08}{62.08}{64.08}{red}{white}{blue}{50}}

\newcommand{\aroar}[1]{\gradientcelld{#1}{77.67}{79.67}{81.67}{red}{white}{blue}{50}}
\newcommand{\aroen}[1]{\gradientcelld{#1}{72.67}{74.67}{76.67}{red}{white}{blue}{50}}

\newcommand{\zhozh}[1]{\gradientcelld{#1}{55.67}{57.67}{59.67}{red}{white}{blue}{50}}
\newcommand{\zhoen}[1]{\gradientcelld{#1}{50.8}{52.8}{54.8}{red}{white}{blue}{50}}

\newcommand{\kollllko}[1]{\gradientcelld{#1}{63.53}{65.53}{67.53}{red}{white}{blue}{50}}
\newcommand{\kollllen}[1]{\gradientcelld{#1}{60.78}{62.78}{64.78}{red}{white}{blue}{50}}

\newcommand{\arllllar}[1]{\gradientcelld{#1}{69.42}{71.42}{73.42}{red}{white}{blue}{50}}
\newcommand{\arllllen}[1]{\gradientcelld{#1}{73.49}{75.49}{77.49}{red}{white}{blue}{50}}

\newcommand{\zhllllzh}[1]{\gradientcelld{#1}{61.9}{63.9}{65.9}{red}{white}{blue}{50}}
\newcommand{\zhllllen}[1]{\gradientcelld{#1}{57.09}{59.09}{61.09}{red}{white}{blue}{50}}

\newcommand{\arlar}[1]{\gradientcelld{#1}{69.46}{71.46}{73.46}{red}{white}{blue}{50}}
\newcommand{\arlen}[1]{\gradientcelld{#1}{71.55}{73.55}{75.55}{red}{white}{blue}{50}}

\newcommand{\zhmzh}[1]{\gradientcelld{#1}{58.47}{60.47}{62.47}{red}{white}{blue}{50}}
\newcommand{\zhmen}[1]{\gradientcelld{#1}{52.47}{54.47}{56.47}{red}{white}{blue}{50}}

\newcommand{\komko}[1]{\gradientcelld{#1}{61.57}{63.57}{65.57}{red}{white}{blue}{50}}
\newcommand{\komen}[1]{\gradientcelld{#1}{58.25}{60.25}{62.25}{red}{white}{blue}{50}}

\newcommand{\armar}[1]{\gradientcelld{#1}{59.62}{61.62}{63.62}{red}{white}{blue}{50}}
\newcommand{\armen}[1]{\gradientcelld{#1}{64.48}{66.48}{68.48}{red}{white}{blue}{50}}

\newcommand{\zhgezh}[1]{\gradientcelld{#1}{69.67}{71.67}{73.67}{red}{white}{blue}{50}}
\newcommand{\zhgeen}[1]{\gradientcelld{#1}{59.23}{61.23}{63.23}{red}{white}{blue}{50}}

\newcommand{\kogeko}[1]{\gradientcelld{#1}{76.07}{78.07}{80.07}{red}{white}{blue}{50}}
\newcommand{\kogeen}[1]{\gradientcelld{#1}{65.63}{67.63}{69.63}{red}{white}{blue}{50}}

\newcommand{\argear}[1]{\gradientcelld{#1}{81.76}{83.76}{85.76}{red}{white}{blue}{50}}
\newcommand{\argeen}[1]{\gradientcelld{#1}{75.48}{77.49}{79.49}{red}{white}{blue}{50}}

\newcommand{\langbox}[2]{%
  \tikz[baseline=(char.base)]{
    \node[shape=rectangle, fill=#1, text=white, rounded corners=1pt,  minimum width=1.8em,
    minimum height=1.1em,
    inner sep=0pt] (char) {\footnotesize #2};
  }
}

\newcommand{\bodybox}[2]{%
  \tikz[baseline=(char.base)]{
    \node[shape=rectangle, fill=#1, text=white, rounded corners=1pt,  minimum width=1.4em,
    minimum height=1.1em,
    inner sep=0pt] (char) {\footnotesize #2};
  }
}

%% file: tables/ITO.tex
\begin{table*}[t!]
\caption{Prompt configurations for exploring how language alignment affects factual knowledge-based responses. The full prompt example can be found in the Appendix~\ref{app:prompt}.}\label{tab:prompt}
     \centering
\begin{tabular}{l|ccc|l}
\toprule
 &
  \multicolumn{3}{c|}{\textbf{Language Setting}} &
  \multicolumn{1}{c}{\textbf{Prompt Template}} \\ \hline
\textbf{Prompt Type} &
  \multicolumn{1}{c|}{\textbf{Input}} &
  \multicolumn{1}{c|}{\textbf{Thought}} &
  \textbf{Output} &
  \multicolumn{1}{c}{If \langbox{gray}{LC1} is \bodybox{gray}{EN} and \langbox{orange}{LC2} is \bodybox{orange}{ZH}} \\ \midrule
\textbf{Baseline} &
  \multicolumn{1}{c|}{\langbox{gray}{LC1}} &
  \multicolumn{1}{c|}{-} &
  - &
  \multicolumn{1}{c}{\{Question\}} \\
\textbf{L2T-Consistent} &
  \multicolumn{1}{c|}{\langbox{gray}{LC1}} &
  \multicolumn{1}{c|}{\langbox{gray}{LC1}} &
  - &
  \multicolumn{1}{c}{\textit{Think in English.} \{Question\}} \\
\textbf{L2T-Transfer} &
  \multicolumn{1}{c|}{\langbox{gray}{LC1}} &
  \multicolumn{1}{c|}{\langbox{orange}{LC2}} &
  - &
  \multicolumn{1}{c}{\textit{Think in Chinese.} \{Question\}} \\
\textbf{L2T-Align} &
  \multicolumn{1}{c|}{\langbox{gray}{LC1}} &
  \multicolumn{1}{c|}{\langbox{orange}{LC2}} &
  \multicolumn{1}{c|}{\langbox{gray}{LC1}} &
  \multicolumn{1}{c}{\textit{Think in Chinese and answer in English.} \{Question\}} \\ \bottomrule
\end{tabular}

\end{table*}

%% file: tables/baseline_final.tex
\begin{table}[t!]
    \caption{Accuracy of baseline prompts across input languages. Bold font shows higher accuracy per input language pair.}\label{tab:baseline}
    \centering
\begin{tabular}{l|cc|cc|cc}
\toprule
\multicolumn{1}{c|}{\textbf{Task}} &
  \multicolumn{2}{c|}{\textbf{CMMLU + SE}} &
  \multicolumn{2}{c|}{\textbf{CLIcK}} &
  \multicolumn{2}{c}{\textbf{ArabicMMLU}} \\ \hline
\multicolumn{1}{c|}{\textbf{Input}} &
  \textbf{ZH } &
  \textbf{EN} &
  \textbf{KO} &
  \textbf{EN} &
  \textbf{AR} &
  \textbf{EN} \\ \midrule
\textbf{Mistral}  & \textbf{60.47} & 54.47 & \textbf{63.57} & 60.25 & 61.62 & \textbf{66.48} \\
\textbf{Llama}   & \textbf{63.90} & 59.09 & \textbf{65.53} & 62.78 & 71.42 & \textbf{75.49} \\
\textbf{GPT}     & \textbf{57.67} & 52.80 & \textbf{69.62} & 62.08 & \textbf{79.85} & 74.67 \\
\textbf{Gemini}  & \textbf{71.67} & 61.23 & \textbf{78.07} & 67.63 & \textbf{83.76} & 77.49 \\ \bottomrule
\end{tabular}
\end{table}

%% file: tables/I_T_acc_final.tex
\begin{table}[t!]
\caption{Accuracy under the Input–Thought (I–T) setting. Color intensity reflects gains (blue) or losses (red) relative to the baseline in Table~\ref{tab:baseline}.}\label{fig:I_T_Acc}

\centering
\begin{subtable}[t!]{0.48\textwidth}
    \caption{L2T-Consistent.}\label{tab:Consistent_Acc}
    \centering
    \resizebox{\textwidth}{!}{
        \begin{tabular}{l|cccc|cccc|cccc}
    \toprule
    \multicolumn{1}{c|}{\textbf{Task}} &
      \multicolumn{4}{c|}{\textbf{CMMLU + SE}} &
      \multicolumn{4}{c|}{\textbf{CLIcK}} &
      \multicolumn{4}{c}{\textbf{ArabicMMLU}} \\
 \hline
    \multicolumn{1}{c|}{\textbf{I-T}} &
  \multicolumn{2}{c}{\parbox[c][1.5em][c]{0.16\textwidth}{\centering \textbf{ZH-ZH}}} &
\multicolumn{2}{c|}{\parbox[c][1.5em][c]{0.16\textwidth}{\centering \textbf{EN-EN}}} &
\multicolumn{2}{c}{\parbox[c][1.5em][c]{0.16\textwidth}{\centering \textbf{\mbox{KO-KO}}}} &
\multicolumn{2}{c|}{\parbox[c][1.5em][c]{0.16\textwidth}{\centering \textbf{EN-EN}}} &
\multicolumn{2}{c}{\parbox[c][1.5em][c]{0.16\textwidth}{\centering \textbf{AR-AR}}} &
\multicolumn{2}{c}{\parbox[c][1.5em][c]{0.16\textwidth}{\centering \textbf{EN-EN}}} \\ \midrule
    \textbf{Mistral} & \multicolumn{2}{c}{\zhmzh{60.32}} & \multicolumn{2}{c|}{\zhmen{55.09}} & \multicolumn{2}{c}{\komko{63.74}} & \multicolumn{2}{c|}{\komen{60.10}} & \multicolumn{2}{c}{\armar{61.35}} & \multicolumn{2}{c}{\armen{67.12}} \\ 
    \textbf{Llama} & \multicolumn{2}{c}{\zhllllzh{62.55}} & \multicolumn{2}{c|}{\zhllllen{58.87}} & \multicolumn{2}{c}{\kollllko{64.68}} & \multicolumn{2}{c|}{\kollllen{63.32}} & \multicolumn{2}{c}{\arllllar{71.21}} & \multicolumn{2}{c}{\arllllen{75.28}} \\ 
    \textbf{GPT} & \multicolumn{2}{c}{\zhozh{57.67}} & \multicolumn{2}{c|}{\zhoen{52.8}} & \multicolumn{2}{c}{\kooko{69.47}} & \multicolumn{2}{c|}{\kooen{62.18}} & \multicolumn{2}{c}{\aroar{79.49}} & \multicolumn{2}{c}{\aroen{74.6}} \\ 
    \textbf{Gemini} & \multicolumn{2}{c}{\zhgezh{71.64}} & \multicolumn{2}{c|}{\zhgeen{61.23}} & \multicolumn{2}{c}{\kogeko{77.62}} & \multicolumn{2}{c|}{\kogeen{67.51}} & \multicolumn{2}{c}{\argear{83.95}} & \multicolumn{2}{c}{\argeen{77.6}} \\ \bottomrule
    \end{tabular}
    }
    
\end{subtable}
\begin{subtable}[t!]{0.48\textwidth}
\caption{L2T-Transfer.}
    \centering
    \resizebox{\textwidth}{!}{
    \begin{tabular}{l|cccc|cccc|cccc}
    \toprule
    \multicolumn{1}{c|}{\textbf{Task}} &
      \multicolumn{4}{c|}{\textbf{CMMLU + SE}} &
      \multicolumn{4}{c|}{\textbf{CLIcK}} &
      \multicolumn{4}{c}{\textbf{ArabicMMLU}} \\ \hline
    \multicolumn{1}{c|}{\textbf{I-T}} &
  \multicolumn{2}{c}{\parbox[c][1.5em][c]{0.16\textwidth}{\centering \textbf{ZH-EN}}} &
\multicolumn{2}{c|}{\parbox[c][1.5em][c]{0.16\textwidth}{\centering \textbf{EN-ZH}}} &
\multicolumn{2}{c}{\parbox[c][1.5em][c]{0.16\textwidth}{\centering \textbf{\mbox{KO-EN}}}} &
\multicolumn{2}{c|}{\parbox[c][1.5em][c]{0.16\textwidth}{\centering \textbf{EN-KO}}} &
\multicolumn{2}{c}{\parbox[c][1.5em][c]{0.16\textwidth}{\centering \textbf{AR-EN}}} &
\multicolumn{2}{c}{\parbox[c][1.5em][c]{0.16\textwidth}{\centering \textbf{EN-AR}}} \\  \midrule
    \textbf{Mistral} & \multicolumn{2}{c}{\zhmzh{58.99}} & \multicolumn{2}{c|}{\zhmen{55.13}} & \multicolumn{2}{c}{\komko{62.65}} & \multicolumn{2}{c|}{\komen{60.72}} & \multicolumn{2}{c}{\armar{60.78}} & \multicolumn{2}{c}{\armen{67.78}} \\
    \textbf{Llama} & \multicolumn{2}{c}{\zhllllzh{63.68}} & \multicolumn{2}{c|}{\zhllllen{57.52}} & \multicolumn{2}{c}{\kollllko{65.11}} & \multicolumn{2}{c|}{\kollllen{58.96}} & \multicolumn{2}{c}{\arlar{70.28}} & \multicolumn{2}{c}{\arlen{73.35}} \\
    \textbf{GPT} & \multicolumn{2}{c}{\zhozh{56.95}} & \multicolumn{2}{c|}{\zhoen{52.89}} & \multicolumn{2}{c}{\kooko{68.9}} & \multicolumn{2}{c|}{\kooen{62.73}} & \multicolumn{2}{c}{\aroar{78.94}} & \multicolumn{2}{c}{\aroen{73.76}} \\
    \textbf{Gemini} & \multicolumn{2}{c}{\zhgezh{70.19}} & \multicolumn{2}{c|}{\zhgeen{64.37}} & \multicolumn{2}{c}{\kogeko{75.19}} & \multicolumn{2}{c|}{\kogeen{68.85}} & \multicolumn{2}{c}{\argear{82.4}} & \multicolumn{2}{c}{\argeen{79.13}} \\ \bottomrule
    \end{tabular}
    }
    \label{tab:Transfer_Acc}
\end{subtable}
\end{table}

%% file: images/perplexity.tex
\begin{figure*}[t]
    \centering
    \begin{subfigure}{0.16\textwidth}
        \centering
        \includegraphics[width=\linewidth]{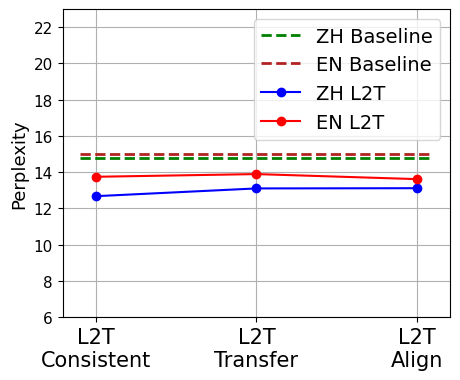}
        \caption{ZH: O, EN: X \textit{(36)}}
        \label{fig:sub5}
    \end{subfigure}
    \begin{subfigure}{0.16\textwidth}
        \centering
        \includegraphics[width=\linewidth]{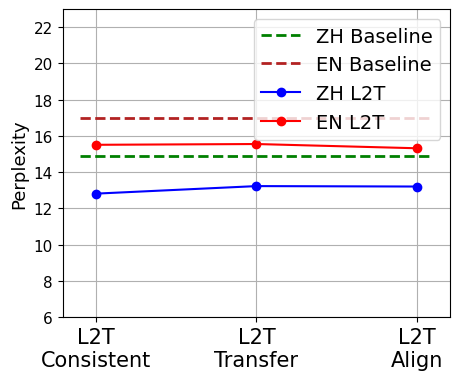}
        \caption{ZH: X, EN: O \textit{(24)}}
        \label{fig:sub6}
    \end{subfigure}
    \begin{subfigure}{0.16\textwidth}
        \centering
        \includegraphics[width=\linewidth]{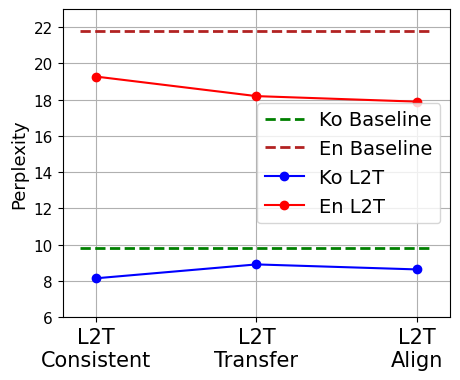}
        \caption{KO: O, EN: X \textit{(51)}}
        \label{fig:sub1}
    \end{subfigure}
    \begin{subfigure}{0.16\textwidth}
        \centering
        \includegraphics[width=\linewidth]{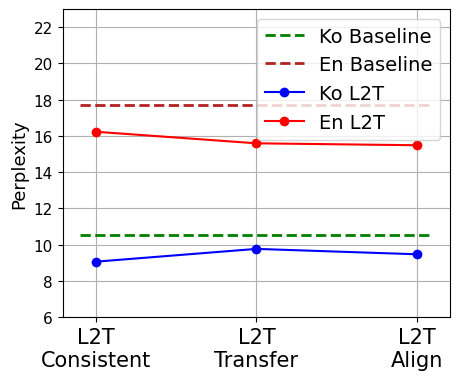}
        \caption{KO: X, EN: O \textit{(27)}}
        \label{fig:sub2}
    \end{subfigure}
    \begin{subfigure}{0.16\textwidth}
        \centering
        \includegraphics[width=\linewidth]{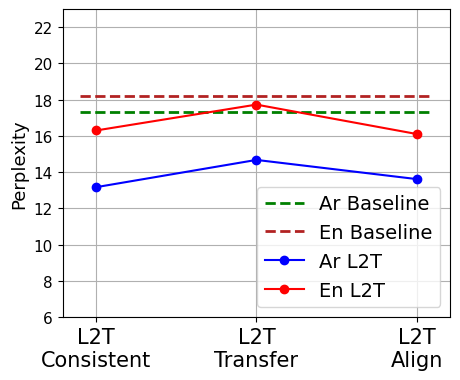}
        \caption{AR: O, EN: X \textit{(42)}}
        \label{fig:sub3}
    \end{subfigure}
    \begin{subfigure}{0.16\textwidth}
        \centering
        \includegraphics[width=\linewidth]{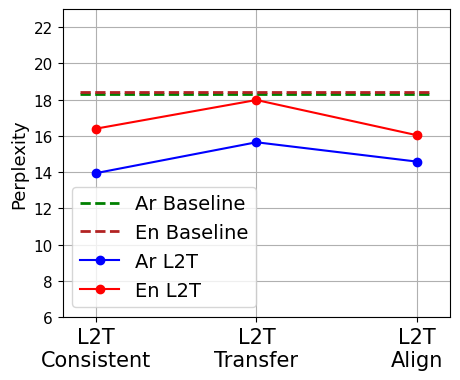}
        \caption{AR: X, EN: O \textit{(95)}}
        \label{fig:sub4}
    \end{subfigure}
    \caption{Average PPL on the Llama. \textit{(numbers)} indicate the number of evaluation samples.}
    \Description{Six line plots comparing the average perplexity of different models across various datasets for the Llama model. Each subfigure (a) to (f) corresponds to a different language pair configuration. The x-axis shows three stages: L2T-Consistent, L2T-Transfer, and L2T-Align. Each plot contains lines for language baselines and L2T approaches.}
    \label{fig:perplexity}
\end{figure*}

%% file: tables/I_T_IN_final.tex
\begin{table}[t!]
\caption{Ratio of responses in the input language. Darker green indicates more responses not in the input language.}
\label{tab:I_T_IN}
\centering
\begin{subtable}[t]{0.48\textwidth}
\caption{Baseline and L2T-Consistent.}
    \label{tab:Consistent_In}
    \centering
    \resizebox{\textwidth}{!}{
    \begin{tabular}{l|cccc|cccc|cccc}
\toprule
\multicolumn{1}{c|}{\textbf{Task}} &
  \multicolumn{4}{c|}{\textbf{CMMLU + SE}} &
  \multicolumn{4}{c|}{\textbf{CLIcK}} &
  \multicolumn{4}{c}{\textbf{ArabicMMLU}} \\ \hline
\multicolumn{1}{c|}{\textbf{Prompt}} &
  \multicolumn{2}{c|}{\makebox[0.16\textwidth][c]{\centering \textbf{Baseline}}} &
  \multicolumn{2}{c|}{\makebox[0.16\textwidth][c]{\centering \textbf{\mbox{L2T-Consistent}}}} &
  \multicolumn{2}{c|}{\makebox[0.16\textwidth][c]{\centering \textbf{Baseline}}} &
  \multicolumn{2}{c|}{\makebox[0.16\textwidth][c]{\centering \textbf{\mbox{L2T-Consistent}}}} &
  \multicolumn{2}{c|}{\makebox[0.16\textwidth][c]{\centering \textbf{Baseline}}} &
  \multicolumn{2}{c}{\makebox[0.16\textwidth][c]{\centering \textbf{\mbox{L2T-Consistent}}}} \\ \hline
\multicolumn{1}{c|}{\textbf{Input}} &
  \textbf{ZH} &
  \multicolumn{1}{c|}{\textbf{EN}} &
  \textbf{ZH} &
  \textbf{EN} &
  \textbf{KO} &
  \multicolumn{1}{c|}{\textbf{EN}} &
  \textbf{KO} &
  \textbf{EN} &
  \textbf{AR} &
  \multicolumn{1}{c|}{\textbf{EN}} &
  \textbf{AR} &
  \textbf{EN} \\ \midrule
\textbf{Mistral} &
  \relang{99.85} &
  \multicolumn{1}{c|}{\relang{100}} &
  \relang{99.84} &
  \relang{100} &
  \relang{99.95} &
  \multicolumn{1}{c|}{\relang{100}} &
  \relang{100} &
  \relang{100} &
  \relang{99.91} &
  \multicolumn{1}{c|}{\relang{99.91}} &
  \relang{99.98} &
  \relang{99.88} \\
\textbf{Llama} &
  \relang{97.11} &
  \multicolumn{1}{c|}{\relang{100}} &
  \relang{98.46} &
  \relang{100} &
  \relang{99.95} &
  \multicolumn{1}{c|}{\relang{100}} &
  \relang{99.78} &
  \relang{99.98} &
  \relang{99.77} &
  \multicolumn{1}{c|}{\relang{99.98}} &
  \relang{99.91} &
  \relang{100} \\
\textbf{GPT} &
  \relang{99.97} &
  \multicolumn{1}{c|}{\relang{100}} &
  \relang{100} &
  \relang{100} &
  \relang{100} &
  \multicolumn{1}{c|}{\relang{100}} &
  \relang{100} &
  \relang{100} &
  \relang{100} &
  \multicolumn{1}{c|}{\relang{99.98}} &
  \relang{100} &
  \relang{100} \\
\textbf{Gemini} &
  \relang{99.81} &
  \multicolumn{1}{c|}{\relang{100}} &
  \relang{99.78} &
  \relang{99.97} &
  \relang{99.98} &
  \multicolumn{1}{c|}{\relang{99.98}} &
  \relang{99.98} &
  \relang{100} &
  \relang{99.98} &
  \multicolumn{1}{c|}{\relang{100}} &
  \relang{100} &
  \relang{99.95} \\ \bottomrule
\end{tabular}
    }
    
\end{subtable}
\begin{subtable}[t]{0.48\textwidth}
\caption{L2T-Transfer and L2T-Align.}
    \label{tab:Transfer_In}
    \centering
    \resizebox{\textwidth}{!}{
    \begin{tabular}{l|cccc|cccc|cccc}
\toprule
\multicolumn{1}{c|}{\textbf{Task}} &
  \multicolumn{4}{c|}{\textbf{CMMLU + SE}} &
  \multicolumn{4}{c|}{\textbf{CLIcK}} &
  \multicolumn{4}{c}{\textbf{ArabicMMLU}} \\ \hline
\multicolumn{1}{c|}{\textbf{Prompt}} &
  \multicolumn{2}{c|}{\makebox[0.16\textwidth][c]{\centering \textbf{L2T-Transfer}}} &
  \multicolumn{2}{c|}{\makebox[0.16\textwidth][c]{\centering \textbf{L2T-Align}}} &
  \multicolumn{2}{c|}{\makebox[0.16\textwidth][c]{\centering \textbf{L2T-Transfer}}} &
  \multicolumn{2}{c|}{\makebox[0.16\textwidth][c]{\centering \textbf{L2T-Transfer}}} &
  \multicolumn{2}{c|}{\makebox[0.16\textwidth][c]{\centering \textbf{L2T-Align}}} &
  \multicolumn{2}{c}{\makebox[0.16\textwidth][c]{\centering \textbf{L2T-Align}}} \\ \hline
\multicolumn{1}{c|}{\textbf{Thought}} &
  \textbf{EN} &
  \multicolumn{1}{c|}{\textbf{ZH}} &
  \textbf{EN} &
  \textbf{ZH} &
  \textbf{EN} &
  \multicolumn{1}{c|}{\textbf{KO}} &
  \textbf{EN} &
  \textbf{KO} &
  \textbf{EN} &
  \multicolumn{1}{c|}{\textbf{AR}} &
  \textbf{EN} &
  \textbf{AR} \\ \midrule
\textbf{Mistral} &
  \relang{97.96} &
  \multicolumn{1}{c|}{\relang{99.97}} &
  \relang{99.88} &
  \relang{100} &
  \relang{76.11} &
  \multicolumn{1}{c|}{\relang{100}} &
  \relang{99.9} &
  \relang{100} &
  \relang{99.93} &
  \multicolumn{1}{c|}{\relang{99.75}} &
  \relang{99.95} &
  \relang{99.77} \\
\textbf{Llama} &
  \relang{89.78} &
  \multicolumn{1}{c|}{\relang{41.48}} &
  \relang{98.77} &
  \relang{99.97} &
  \relang{99.95} &
  \multicolumn{1}{c|}{\relang{67.95}} &
  \relang{99.78} &
  \relang{100} &
  \relang{8.66} &
  \multicolumn{1}{c|}{\relang{79.17}} &
  \relang{99.95} &
  \relang{99.66} \\
\textbf{GPT} &
  \relang{99.91} &
  \multicolumn{1}{c|}{\relang{3.05}} &
  \relang{99.97} &
  \relang{99.97} &
  \relang{99.98} &
  \multicolumn{1}{c|}{\relang{57.89}} &
  \relang{100} &
  \relang{100} &
  \relang{99.91} &
  \multicolumn{1}{c|}{\relang{52.46}} &
  \relang{99.98} &
  \relang{99.95} \\
\textbf{Gemini} &
  \relang{99.37} &
  \multicolumn{1}{c|}{\relang{5.22}} &
  \relang{99.87} &
  \relang{27.52} &
  \relang{43.44} &
  \multicolumn{1}{c|}{\relang{1.46}} &
  \relang{93.88} &
  \relang{99.3} &
  \relang{96.54} &
  \multicolumn{1}{c|}{\relang{0}} &
  \relang{100} &
  \relang{99.82} \\ \bottomrule
\end{tabular}
    }
    
\end{subtable}
\end{table}

%% file: tables/equal.tex
\begin{table}[t!]
\caption{Accuracy under L2T-Align. Performance improves when internal thought occurs in the original language and output matches the input (blue), but drops when thought is in English (red) relative to the baseline.}\label{tab:equal}
    \centering
    \resizebox{\columnwidth}{!}{
\begin{tabular}{l|cc|cc|cc}
\toprule
\multicolumn{1}{c|}{\textbf{Task}} &
  \multicolumn{2}{c|}{\textbf{CMMLU + SE}} &
  \multicolumn{2}{c|}{\textbf{CLIcK}} &
  \multicolumn{2}{c}{\textbf{ArabicMMLU}} \\ \hline
\multicolumn{1}{c|}{\textbf{Thought}} &
  \textbf{EN} &
  \textbf{ZH} &
  \textbf{EN} &
  \textbf{KO} &
  \textbf{EN} &
  \textbf{AR} \\ \midrule
\textbf{Mistral} &
  \zhmzh{59.75} &
  \zhmen{55.79} &
  \komko{63.02} &
  \komen{60.94} &
  \armar{60.57} &
  \armen{69.17} \\
\textbf{Llama} &
  \zhllllzh{63.68} &
  \zhllllen{60.75} &
  \kollllko{64.83} &
  \kollllen{64.49} &
  \arllllar{71.08} &
  \arllllen{75.63} \\
\textbf{GPT} &
  \zhozh{57.36} &
  \zhoen{53.08} &
  \kooko{69.07} &
  \kooen{63.15} &
  \aroar{78.83} &
  \aroen{75.53} \\
\textbf{Gemini} &
  \zhgezh{70.0} &
  \zhgeen{64.4} &
  \kogeko{75.32} &
  \kogeen{69.27} &
  \argear{83.04} &
  \argeen{78.24} \\ \bottomrule
\end{tabular}
    }
\end{table}

%% file: reference.bbl

%% file: tables/model_analysis.tex
\begin{table*}[t!]
\centering
 \resizebox{\textwidth}{!}{%
\begin{tabular}{l|c|c|c|c|c|c}
\hline
                           & \textbf{MMLU} & \textbf{MMLU-pro} & \textbf{MMMLU-all} & \textbf{MMMLU-ZH} & \textbf{MMMLU-KO} & \textbf{MMMLU-AR} \\ \hline
\textbf{Mistral-Small-24B} \cite{mistral} & 81            & -                 & -                  & \( \approx \)70               & \( \approx \)55               & -                 \\
\textbf{Llama3.1-70B} \cite{Llama3_1}      & 84          & 66              & -                  & -                 & -                 & -                 \\
\textbf{GPT-4o-mini} \cite{gpt4o}       & 82            & -                 & 59               & 73                & 72                & 71                \\
\textbf{Gemini-2.0-Lite} \cite{gemini_lite}  & -             & 72              & 78               & -                 & -                 & -                 \\ \hline
\end{tabular}
}
\caption{Official performance of each model on the MMLU dataset and its multilingual variants.}
\label{tab:model_analysis}
\end{table*}

%% file: tables/baseline_appendix.tex
\begin{table*}[ht!]
    \centering
    \resizebox{\textwidth}{!}{
\begin{tabular}{l|cccc|cccc|cccc}
\hline
\multicolumn{1}{c|}{\textbf{Task}} &
  \multicolumn{4}{c|}{\textbf{Chinese-Specific}} &
  \multicolumn{4}{c|}{\textbf{Korean-Specific}} &
  \multicolumn{4}{c}{\textbf{Arabic-Specific}} \\ \hline
\multicolumn{1}{c|}{\textbf{Input}} &
  \multicolumn{2}{c|}{\textbf{ZH}} &
  \multicolumn{2}{c|}{\textbf{EN}} &
  \multicolumn{2}{c|}{\textbf{KO}} &
  \multicolumn{2}{c|}{\textbf{EN}} &
  \multicolumn{2}{c|}{\textbf{AR}} &
  \multicolumn{2}{c}{\textbf{EN}} \\ \hline
\multicolumn{1}{c|}{\textbf{}} &
  \multicolumn{1}{c|}{\textbf{Acc}} &
  \multicolumn{1}{c|}{\textbf{In \%}} &
  \multicolumn{1}{c|}{\textbf{Acc}} &
  \textbf{IN \%} &
  \multicolumn{1}{c|}{\textbf{Acc}} &
  \multicolumn{1}{c|}{\textbf{In \%}} &
  \multicolumn{1}{c|}{\textbf{Acc}} &
  \textbf{IN \%} &
  \multicolumn{1}{c|}{\textbf{Acc}} &
  \multicolumn{1}{c|}{\textbf{In \%}} &
  \multicolumn{1}{c|}{\textbf{Acc}} &
  \textbf{IN \%} \\ \hline
  \textbf{Mistral-Small-24B}          & \multicolumn{1}{c|}{\textbf{60.47 (±0.77)}} & \multicolumn{1}{c|}{99.85}          & \multicolumn{1}{c|}{54.47 (±1.94)} & 100                                 & \multicolumn{1}{c|}{\textbf{63.57 (±1.0)}}  & \multicolumn{1}{c|}{99.95}          & \multicolumn{1}{c|}{60.25 (±0.69)}          & 100            & \multicolumn{1}{c|}{61.62 (±0.93)}          & \multicolumn{1}{c|}{99.91}          & \multicolumn{1}{c|}{66.48 (±0.8)}           & 99.91          \\ \hline
\textbf{Llama3-8B}                  & \multicolumn{1}{c|}{\textbf{44.78 (±0.11)}} & \multicolumn{1}{c|}{\textit{0.03}}  & \multicolumn{1}{c|}{43.46 (±1.67)} & \multicolumn{1}{c|}{100}            & \multicolumn{1}{c|}{\textbf{51.2 (±1.25)}}  & \multicolumn{1}{c|}{\textit{0.25}}  & \multicolumn{1}{c|}{50.83 (±0.23)}          & 100            & \multicolumn{1}{c|}{48.0 (±0.85)}           & \multicolumn{1}{c|}{99.57}          & \multicolumn{1}{c|}{\textbf{59.5 (±0.99)}}  & 99.95          \\
\textbf{Llama3.1-8B}                & \multicolumn{1}{c|}{\textbf{49.43 (±1.72)}} & \multicolumn{1}{c|}{99.84}          & \multicolumn{1}{c|}{44.5 (±1.43)}  & \multicolumn{1}{c|}{100}            & \multicolumn{1}{c|}{54.62 (±0.21)}          & \multicolumn{1}{c|}{99.93}          & \multicolumn{1}{c|}{\textbf{54.67 (±1.39)}} & 99.9           & \multicolumn{1}{c|}{53.62 (±0.76)}          & \multicolumn{1}{c|}{100}            & \multicolumn{1}{c|}{\textbf{60.78 (±0.54)}} & 99.8           \\ \hline
\textbf{Llama3-70B}                 & \multicolumn{1}{c|}{\textbf{58.24 (±1.32)}} & \multicolumn{1}{c|}{\textit{9.72}}  & \multicolumn{1}{c|}{56.1 (±0.62)}  & \multicolumn{1}{c|}{100}            & \multicolumn{1}{c|}{\textbf{64.19 (±0.11)}} & \multicolumn{1}{c|}{\textit{0.59}}  & \multicolumn{1}{c|}{61.81 (±0.6)}           & 100            & \multicolumn{1}{c|}{71.49 (±0.79)}          & \multicolumn{1}{c|}{99.84}          & \multicolumn{1}{c|}{\textbf{73.35 (±0.9)}}  & 100            \\
\textbf{Llama3.1-70B}               & \multicolumn{1}{c|}{\textbf{63.9 (±2.1)}}   & \multicolumn{1}{c|}{97.11}          & \multicolumn{1}{c|}{59.09 (±0.96)} & \multicolumn{1}{c|}{100}            & \multicolumn{1}{c|}{\textbf{65.53 (±1.32)}} & \multicolumn{1}{c|}{99.95}          & \multicolumn{1}{c|}{62.78 (±0.93)}          & 100            & \multicolumn{1}{c|}{71.42 (±0.79)}          & \multicolumn{1}{c|}{99.77}          & \multicolumn{1}{c|}{\textbf{75.49 (±0.39)}} & 99.98          \\ \hline
\textbf{GPT-3.5-turbo}              & \multicolumn{1}{c|}{\textbf{45.09 (±0.57)}} & \multicolumn{1}{c|}{99.4}           & \multicolumn{1}{c|}{44.34 (±2.22)} & \multicolumn{1}{c|}{99.97}          & \multicolumn{1}{c|}{\textbf{57.22 (±0.5)}}  & \multicolumn{1}{c|}{99.9}           & \multicolumn{1}{c|}{56.51 (±0.65)}          & 100            & \multicolumn{1}{c|}{48.02 (±0.43)}          & \multicolumn{1}{c|}{100}            & \multicolumn{1}{c|}{\textbf{64.96 (±0.22)}} & 100            \\
\textbf{GPT-4o-mini}                & \multicolumn{1}{c|}{\textbf{57.67 (±0.3)}}  & \multicolumn{1}{c|}{99.97}          & \multicolumn{1}{c|}{52.8 (±0.24)}  & \multicolumn{1}{c|}{100}            & \multicolumn{1}{c|}{\textbf{69.62 (±1.08)}} & \multicolumn{1}{c|}{100}            & \multicolumn{1}{c|}{62.08 (±0.34)}          & 100            & \multicolumn{1}{c|}{\textbf{79.85 (±0.51)}} & \multicolumn{1}{c|}{100}            & \multicolumn{1}{c|}{74.67 (±0.28)}          & 99.98          \\ \hline
\textbf{Gemini-2.0-Lite}            & \multicolumn{1}{c|}{\textbf{71.67 (±0.43)}} & \multicolumn{1}{c|}{99.81}          & \multicolumn{1}{c|}{61.23 (±0.16)} & 100                                 & \multicolumn{1}{c|}{\textbf{78.07 (±0.13)}} & \multicolumn{1}{c|}{99.98}          & \multicolumn{1}{c|}{67.63 (±0.43)}          & 99.98          & \multicolumn{1}{c|}{\textbf{83.76 (±0.49)}} & \multicolumn{1}{c|}{99.98}          & \multicolumn{1}{c|}{77.49 (±0.24)}          & 100            \\ \hline
    \end{tabular}
    }
    \caption{Baseline results.}\label{tab:baseline_appendix}
\end{table*}

%% file: tables/mix_same_appendix.tex
\begin{table*}[ht!]
    \centering
    \resizebox{\textwidth}{!}{
\begin{tabular}{l|cccc|cccc|cccc}
\hline
\multicolumn{1}{c|}{\textbf{Task}} &
  \multicolumn{4}{c|}{\textbf{Chinese-Specific}} &
  \multicolumn{4}{c|}{\textbf{Korean-Specific}} &
  \multicolumn{4}{c}{\textbf{Arabic-Specific}} \\ \hline
\multicolumn{1}{c|}{\textbf{Input-Thought}} &
  \multicolumn{2}{c|}{\textbf{ZH-ZH}} &
  \multicolumn{2}{c|}{\textbf{EN-EN}} &
  \multicolumn{2}{c|}{\textbf{KO-KO}} &
  \multicolumn{2}{c|}{\textbf{EN-EN}} &
  \multicolumn{2}{c|}{\textbf{AR-AR}} &
  \multicolumn{2}{c}{\textbf{EN-EN}} \\ \hline
\multicolumn{1}{c|}{\textbf{}} &
  \multicolumn{1}{c|}{\textbf{Acc}} &
  \multicolumn{1}{c|}{\textbf{IN \%}} &
  \multicolumn{1}{c|}{\textbf{Acc}} &
  \textbf{IN \%} &
  \multicolumn{1}{c|}{\textbf{Acc}} &
  \multicolumn{1}{c|}{\textbf{IN \%}} &
  \multicolumn{1}{c|}{\textbf{Acc}} &
  \textbf{IN \%} &
  \multicolumn{1}{c|}{\textbf{Acc}} &
  \multicolumn{1}{c|}{\textbf{IN \%}} &
  \multicolumn{1}{c|}{\textbf{Acc}} &
  \textbf{IN \%} \\ \hline
  \textbf{Mistral-Small-24B}                  & \multicolumn{1}{c|}{60.32 (±0.29)} & \multicolumn{1}{c|}{99.84}          & \multicolumn{1}{c|}{55.09 (±0.77)} & 100            & \multicolumn{1}{c|}{63.74 (±0.62)} & \multicolumn{1}{c|}{100}            & \multicolumn{1}{c|}{60.10 (±0.19)} & 100            & \multicolumn{1}{c|}{61.35 (±0.64)} & \multicolumn{1}{c|}{99.98}          & \multicolumn{1}{c|}{67.12 (±0.36)} & 99.88          \\ \hline
\textbf{Llama3-8B}                          & \multicolumn{1}{c|}{45.97 (±0.58)} & \multicolumn{1}{c|}{\textit{47.17}} & \multicolumn{1}{c|}{43.21 (±0.41)} & 100            & \multicolumn{1}{c|}{54.92 (±0.31)} & \multicolumn{1}{c|}{99.16}          & \multicolumn{1}{c|}{51.82 (±0.32)} & 100            & \multicolumn{1}{c|}{48.29 (±0.38)} & \multicolumn{1}{c|}{100}            & \multicolumn{1}{c|}{59.5 (±1.01)}  & 99.98          \\
\textbf{Llama3.1-8B}                        & \multicolumn{1}{c|}{48.84 (±1.06)} & \multicolumn{1}{c|}{99.94}          & \multicolumn{1}{c|}{44.62 (±1.24)} & 100            & \multicolumn{1}{c|}{54.5 (±1.27)}  & \multicolumn{1}{c|}{99.88}          & \multicolumn{1}{c|}{53.83 (±1.36)} & 99.98          & \multicolumn{1}{c|}{52.96 (±0.28)} & \multicolumn{1}{c|}{99.98}          & \multicolumn{1}{c|}{60.48 (±0.4)}  & 99.86          \\ \hline
\textbf{Llama3-70B}                         & \multicolumn{1}{c|}{57.17 (±0.98)} & \multicolumn{1}{c|}{\textit{72.67}} & \multicolumn{1}{c|}{54.53 (±0.57)} & 100            & \multicolumn{1}{c|}{63.25 (±0.26)} & \multicolumn{1}{c|}{\textit{36.16}} & \multicolumn{1}{c|}{61.34 (±0.78)} & 100            & \multicolumn{1}{c|}{71.28 (±0.62)} & \multicolumn{1}{c|}{99.89}          & \multicolumn{1}{c|}{72.76 (±0.48)} & 100            \\
\textbf{Llama3.1-70B}                       & \multicolumn{1}{c|}{62.55 (±1.3)}  & \multicolumn{1}{c|}{98.46}          & \multicolumn{1}{c|}{58.87 (±2.07)} & 100            & \multicolumn{1}{c|}{64.68 (±0.34)} & \multicolumn{1}{c|}{99.78}          & \multicolumn{1}{c|}{63.32 (±0.48)} & 99.98          & \multicolumn{1}{c|}{71.21 (±0.48)} & \multicolumn{1}{c|}{99.91}          & \multicolumn{1}{c|}{75.28 (±0.58)} & 100            \\ \hline
\textbf{GPT-3.5-turbo}                      & \multicolumn{1}{c|}{44.53 (±1.7)}  & \multicolumn{1}{c|}{99.78}          & \multicolumn{1}{c|}{44.56 (±0.52)} & 99.97          & \multicolumn{1}{c|}{56.31 (±0.37)} & \multicolumn{1}{c|}{99.9}           & \multicolumn{1}{c|}{57.67 (±0.91)} & 100            & \multicolumn{1}{c|}{53.71 (±1.45)} & \multicolumn{1}{c|}{99.98}          & \multicolumn{1}{c|}{64.55 (±0.37)} & 99.98          \\
\textbf{GPT-4o-mini}                        & \multicolumn{1}{c|}{57.67 (±0.55)} & \multicolumn{1}{c|}{100}            & \multicolumn{1}{c|}{52.8 (±0.33)}  & 100            & \multicolumn{1}{c|}{69.47 (±0.45)} & \multicolumn{1}{c|}{100}            & \multicolumn{1}{c|}{62.18 (±0.24)} & 100            & \multicolumn{1}{c|}{79.49 (±0.25)} & \multicolumn{1}{c|}{100}            & \multicolumn{1}{c|}{74.6 (±0.66)}  & 100            \\ \hline
\textbf{Gemini-2.0-Lite}                    & \multicolumn{1}{c|}{71.64 (±0.63)} & \multicolumn{1}{c|}{99.78}          & \multicolumn{1}{c|}{61.23 (±1.16)} & 99.97          & \multicolumn{1}{c|}{77.62 (±0.45)} & \multicolumn{1}{c|}{99.98}          & \multicolumn{1}{c|}{67.51 (±0.88)} & 100            & \multicolumn{1}{c|}{83.95 (±0.2)}  & \multicolumn{1}{c|}{100}            & \multicolumn{1}{c|}{77.6 (±1.15)}  & 99.95          \\ \hline
    \end{tabular}
    }
    \caption{L2T-Consistent results.}\label{tab:mix_same_appendix}
\end{table*}

%% file: tables/mix_different_appendix.tex
\begin{table*}[ht!]
    \centering
    \resizebox{\textwidth}{!}{
\begin{tabular}{l|cccc|cccc|cccc}
\hline
\multicolumn{1}{c|}{\textbf{Task}} &
  \multicolumn{4}{c|}{\textbf{Chinese-Specific}} &
  \multicolumn{4}{c|}{\textbf{Korean-Specific}} &
  \multicolumn{4}{c}{\textbf{Arabic-Specific}} \\ \hline
\multicolumn{1}{c|}{\textbf{Input-Thought}} &
  \multicolumn{2}{c|}{\textbf{ZH-EN}} &
  \multicolumn{2}{c|}{\textbf{EN-ZH}} &
  \multicolumn{2}{c|}{\textbf{KO-EN}} &
  \multicolumn{2}{c|}{\textbf{EN-KO}} &
  \multicolumn{2}{c|}{\textbf{AR-EN}} &
  \multicolumn{2}{c}{\textbf{EN-AR}} \\ \hline
\multicolumn{1}{c|}{\textbf{}} &
  \multicolumn{1}{c|}{\textbf{Acc}} &
  \multicolumn{1}{c|}{\textbf{IN \%}} &
  \multicolumn{1}{c|}{\textbf{Acc}} &
  \textbf{IN \%} &
  \multicolumn{1}{c|}{\textbf{Acc}} &
  \multicolumn{1}{c|}{\textbf{IN \%}} &
  \multicolumn{1}{c|}{\textbf{Acc}} &
  \textbf{IN \%} &
  \multicolumn{1}{c|}{\textbf{Acc}} &
  \multicolumn{1}{c|}{\textbf{IN \%}} &
  \multicolumn{1}{c|}{\textbf{Acc}} &
  \textbf{IN \%} \\ \hline
\textbf{Mistral-Small-24B}                  & \multicolumn{1}{c|}{58.99 (±1.14)} & \multicolumn{1}{c|}{97.96}          & \multicolumn{1}{c|}{55.13 (±0.30)} & 99.97          & \multicolumn{1}{c|}{62.65 (±0.93)} & \multicolumn{1}{c|}{\textit{76.11}} & \multicolumn{1}{c|}{60.72 (±0.55)} & 100            & \multicolumn{1}{c|}{60.78 (±0.1.61)}  & \multicolumn{1}{c|}{99.93}          & \multicolumn{1}{c|}{67.78 (±0.5)}  & 99.75          \\ \hline
\textbf{Llama3-8B}                          & \multicolumn{1}{c|}{44.84 (±1.8)}  & \multicolumn{1}{c|}{\textit{0.03}}  & \multicolumn{1}{c|}{43.68 (±0.75)} & 99.97          & \multicolumn{1}{c|}{53.41 (±1.25)} & \multicolumn{1}{c|}{\textit{0.07}}  & \multicolumn{1}{c|}{53.63 (±0.67)} & 99.98          & \multicolumn{1}{c|}{49.16 (±0.29)} & \multicolumn{1}{c|}{97.84}          & \multicolumn{1}{c|}{59.21 (±0.72)} & 99.95          \\
\textbf{Llama3.1-8B}                        & \multicolumn{1}{c|}{48.49 (±1.16)} & \multicolumn{1}{c|}{99.97}          & \multicolumn{1}{c|}{45.57 (±0.29)} & 100            & \multicolumn{1}{c|}{54.55 (±0.49)} & \multicolumn{1}{c|}{99.9}           & \multicolumn{1}{c|}{55.02 (±0.73)} & 99.63          & \multicolumn{1}{c|}{53.55 (±0.6)}  & \multicolumn{1}{c|}{99.98}          & \multicolumn{1}{c|}{53.77 (±1.13)} & \textit{44.02} \\ \hline
\textbf{Llama3-70B}                         & \multicolumn{1}{c|}{57.14 (±0.14)} & \multicolumn{1}{c|}{\textit{0}}     & \multicolumn{1}{c|}{56.35 (±1.04)} & 100            & \multicolumn{1}{c|}{63.25 (±0.64)} & \multicolumn{1}{c|}{\textit{0}}     & \multicolumn{1}{c|}{62.01 (±0.37)} & 100            & \multicolumn{1}{c|}{70.33 (±0.71)} & \multicolumn{1}{c|}{\textit{7.89}}  & \multicolumn{1}{c|}{72.46 (±0.17)} & 100            \\
\textbf{Llama3.1-70B}                       & \multicolumn{1}{c|}{63.68 (±0.9)}  & \multicolumn{1}{c|}{\textit{89.78}} & \multicolumn{1}{c|}{57.52 (±1.04)} & \textit{41.48} & \multicolumn{1}{c|}{65.11 (±0.57)} & \multicolumn{1}{c|}{99.95}          & \multicolumn{1}{c|}{58.96 (±0.27)} & \textit{67.95} & \multicolumn{1}{c|}{70.28 (±0.81)} & \multicolumn{1}{c|}{\textit{8.66}}  & \multicolumn{1}{c|}{73.35 (±1.06)} & \textit{79.17} \\ \hline
\textbf{GPT-3.5-turbo}                      & \multicolumn{1}{c|}{43.84 (±0.22)} & \multicolumn{1}{c|}{99.72}          & \multicolumn{1}{c|}{42.64 (±0.94)} & \textit{22.07} & \multicolumn{1}{c|}{56.18 (±0.16)} & \multicolumn{1}{c|}{99.95}          & \multicolumn{1}{c|}{54.82 (±1.02)} & \textit{30.01} & \multicolumn{1}{c|}{53.37 (±1.23)} & \multicolumn{1}{c|}{\textit{46.48}} & \multicolumn{1}{c|}{55.39 (±1.07)} & \textit{31.1}  \\
\textbf{GPT-4o-mini}                        & \multicolumn{1}{c|}{56.95 (±0.82)} & \multicolumn{1}{c|}{99.91}          & \multicolumn{1}{c|}{52.89 (±0.29)} & \textit{3.05}  & \multicolumn{1}{c|}{68.9 (±0.31)}  & \multicolumn{1}{c|}{99.98}          & \multicolumn{1}{c|}{62.73 (±0.35)} & \textit{57.89} & \multicolumn{1}{c|}{78.94 (±0.66)} & \multicolumn{1}{c|}{99.91}          & \multicolumn{1}{c|}{73.76 (±0.28)} & \textit{52.46} \\ \hline
\textbf{Gemini-2.0-Lite}                    & \multicolumn{1}{c|}{70.19 (±0.25)} & \multicolumn{1}{c|}{99.37}          & \multicolumn{1}{c|}{64.37 (±0.89)} & \textit{5.22}  & \multicolumn{1}{c|}{75.19 (±0.49)} & \multicolumn{1}{c|}{\textit{43.44}} & \multicolumn{1}{c|}{68.85 (±0.37)} & \textit{1.46}  & \multicolumn{1}{c|}{82.4 (±0.36)}  & \multicolumn{1}{c|}{96.54}          & \multicolumn{1}{c|}{79.13 (±0.48)} & \textit{0}    
\\ \hline
    \end{tabular}
    }
    \caption{L2T-Transfer results.}\label{tab:mix_different_appendix}
\end{table*}

%% file: tables/equal_appendix.tex
\begin{table*}[t!]
    \centering
    \resizebox{\textwidth}{!}{
\begin{tabular}{l|cccc|cccc|cccc}
\hline
\multicolumn{1}{c|}{\textbf{Task}} &
  \multicolumn{4}{c|}{\textbf{Chinese-Specific}} &
  \multicolumn{4}{c|}{\textbf{Korean-Specific}} &
  \multicolumn{4}{c}{\textbf{Arabic-Specific}} \\ \hline
\multicolumn{1}{c|}{\textbf{Input-Think-Output}} &
  \multicolumn{2}{c|}{\textbf{ZH-EN-ZH}} &
  \multicolumn{2}{c|}{\textbf{EN-ZH-EN}} &
  \multicolumn{2}{c|}{\textbf{KO-EN-KO}} &
  \multicolumn{2}{c|}{\textbf{EN-KO-EN}} &
  \multicolumn{2}{c|}{\textbf{AR-EN-AR}} &
  \multicolumn{2}{c}{\textbf{EN-AR-EN}} \\ \hline
\multicolumn{1}{c|}{\textbf{}} &
  \multicolumn{1}{c|}{\textbf{Acc}} &
  \multicolumn{1}{c|}{\textbf{IN \%}} &
  \multicolumn{1}{c|}{\textbf{Acc}} &
  \textbf{IN \%} &
  \multicolumn{1}{c|}{\textbf{Acc}} &
  \multicolumn{1}{c|}{\textbf{IN \%}} &
  \multicolumn{1}{c|}{\textbf{Acc}} &
  \textbf{IN \%} &
  \multicolumn{1}{c|}{\textbf{Acc}} &
  \multicolumn{1}{c|}{\textbf{IN \%}} &
  \multicolumn{1}{c|}{\textbf{Acc}} &
  \textbf{IN \%} \\ \hline
\textbf{Mistral-Small-24B}                       & \multicolumn{1}{c|}{59.75 (±0.72)} & \multicolumn{1}{c|}{99.88}          & \multicolumn{1}{c|}{55.79 (±0.76)} & 100            & \multicolumn{1}{c|}{63.02 (±0.45)} & \multicolumn{1}{c|}{99.9}           & \multicolumn{1}{c|}{60.94 (±0.57)} & 100            & \multicolumn{1}{c|}{60.57 (±0.51)} & \multicolumn{1}{c|}{99.95}          & \multicolumn{1}{c|}{69.17 (±0.58)} & 99.77          \\ \hline
\textbf{Llama3-8B}                               & \multicolumn{1}{c|}{45.5 (±1.04)}  & \multicolumn{1}{c|}{\textit{0.16}}  & \multicolumn{1}{c|}{43.05 (±0.63)} & 99.97          & \multicolumn{1}{c|}{55.22 (±0.76)} & \multicolumn{1}{c|}{\textit{0.64}}  & \multicolumn{1}{c|}{54.3 (±1.06)}  & 99.95          & \multicolumn{1}{c|}{47.61 (±0.67)} & \multicolumn{1}{c|}{100}            & \multicolumn{1}{c|}{59.6 (±0.08)}  & 99.84          \\
\textbf{Llama3.1-8B}                             & \multicolumn{1}{c|}{49.18 (±1.11)} & \multicolumn{1}{c|}{99.94}          & \multicolumn{1}{c|}{45.25 (±0.46)} & 100            & \multicolumn{1}{c|}{55.66 (±0.97)} & \multicolumn{1}{c|}{99.9}           & \multicolumn{1}{c|}{55.79 (±1.14)} & 99.95          & \multicolumn{1}{c|}{52.89 (±0.85)} & \multicolumn{1}{c|}{100}            & \multicolumn{1}{c|}{61.3 (±0.39)}  & 99.66          \\ \hline
\textbf{Llama3-70B}                              & \multicolumn{1}{c|}{58.05 (±1.66)} & \multicolumn{1}{c|}{\textit{0.06}}  & \multicolumn{1}{c|}{56.57 (±1.09)} & 99.94          & \multicolumn{1}{c|}{63.32 (±0.82)} & \multicolumn{1}{c|}{\textit{0.1}}   & \multicolumn{1}{c|}{62.53 (±0.52)} & 99.98          & \multicolumn{1}{c|}{71.01 (±0.36)} & \multicolumn{1}{c|}{99.95}          & \multicolumn{1}{c|}{73.17 (±0.35)} & 99.95          \\
\textbf{Llama3.1-70B}                            & \multicolumn{1}{c|}{63.68 (±0.1)}  & \multicolumn{1}{c|}{98.77}          & \multicolumn{1}{c|}{60.75 (±0.75)} & 99.97          & \multicolumn{1}{c|}{64.83 (±1.35)} & \multicolumn{1}{c|}{99.95}          & \multicolumn{1}{c|}{64.49 (±0.49)} & 100            & \multicolumn{1}{c|}{71.08 (±0.35)} & \multicolumn{1}{c|}{99.95}          & \multicolumn{1}{c|}{75.63 (±0.92)} & 99.66          \\ \hline
\textbf{GPT-3.5-turbo}                           & \multicolumn{1}{c|}{44.69 (±0.8)}  & \multicolumn{1}{c|}{\textit{85.69}} & \multicolumn{1}{c|}{45.47 (±0.38)} & 100            & \multicolumn{1}{c|}{56.33 (±0.58)} & \multicolumn{1}{c|}{99.98}          & \multicolumn{1}{c|}{57.84 (±1.17)} & 99.93          & \multicolumn{1}{c|}{55.43 (±1.78)} & \multicolumn{1}{c|}{99.7}           & \multicolumn{1}{c|}{66.78 (±1.05)} & 99.7           \\
\textbf{GPT-4o-mini}                             & \multicolumn{1}{c|}{57.36 (±1.09)} & \multicolumn{1}{c|}{99.97}          & \multicolumn{1}{c|}{53.08 (±0.52)} & 99.97          & \multicolumn{1}{c|}{69.07 (±0.45)} & \multicolumn{1}{c|}{100}            & \multicolumn{1}{c|}{63.15 (±0.47)} & 100            & \multicolumn{1}{c|}{78.83 (±0.82)} & \multicolumn{1}{c|}{99.98}          & \multicolumn{1}{c|}{75.53 (±0.16)} & 99.95          \\ \hline
\textbf{Gemini-2.0-Lite}                         & \multicolumn{1}{c|}{70.0 (±0.41)}  & \multicolumn{1}{c|}{99.87}          & \multicolumn{1}{c|}{64.4 (±1.01)}  & \textit{27.52}          & \multicolumn{1}{c|}{75.32 (±0.56)} & \multicolumn{1}{c|}{93.88}          & \multicolumn{1}{c|}{69.27 (±0.38)} & 99.3           & \multicolumn{1}{c|}{83.04 (±0.37)} & \multicolumn{1}{c|}{100}            & \multicolumn{1}{c|}{78.24 (±0.14)} & 99.82         
             \\ \hline
    \end{tabular}
    }
    \caption{L2T-Align results.}\label{tab:equal_appendix}
\end{table*}

%% file: tables/persona_prompt.tex
\begin{table*}[t]
  \centering
  \begin{tabular}{l}
    \hline
    \textbf{Persona Instruction} \\
    \hline
    Adopt the identity of \{persona\}. Answer the questions while staying in strict accordance with the nature of this identity. \\
    \hline
    Take the role of \{persona\}. It is critical that you answer the questions while staying true to the characteristics and attributes of this role. \\
    \hline
    You are \{persona\}. Your responses should closely mirror the knowledge and abilities of this persona. \\
    \hline
  \end{tabular}
  \caption{The persona instructions in English.}
  \label{persona_instruction_en}
\end{table*}

\begin{table*}[t]
  \centering
  \begin{tabular}{l}
    \hline
    \textbf{Persona Instruction} \\
    \hline
    \begin{CJK*}{UTF8}{gbsn}采用\{persona\}的身份。严格按照此身份的性质回答问题。\end{CJK*} \\
    \hline
    \begin{CJK*}{UTF8}{gbsn}扮演\{persona\}的角色。回答问题时，务必要忠实于该角色的特征和属性。\end{CJK*} \\
    \hline
    \begin{CJK*}{UTF8}{gbsn}您是\{persona\}。您的回答应该与该角色的知识和能力相符。\end{CJK*} \\
    \hline
  \end{tabular}
  \caption{The persona instructions in Chinese (Simplified).}
  \label{persona_instruction_zh}
\end{table*}

\begin{table*}[t]
  \centering
  \begin{tabular}{l}
    \hline
    \textbf{Persona Instruction} \\
    \hline
    \begin{CJK}{UTF8}{mj}\{persona\}의 정체성을 취해라. 질문에 답할 때 \{persona\}의 특성을 충실하게 따라라.\end{CJK} \\
    \hline
    \begin{CJK}{UTF8}{mj}\{persona\}의 역할을 해라. 질문에 답할 때 \{persona\}의 특성과 속성을 유지하는 것이 중요하다.\end{CJK} \\
    \hline
    \begin{CJK}{UTF8}{mj}당신은 \{persona\}이다. 당신은 \{persona\}의 지식과 능력을 밀접하게 반영하여 응답해야 한다.\end{CJK} \\
    \hline
  \end{tabular}
  \caption{The persona instructions in Korean.}
  \label{persona_instruction_ko}
\end{table*}

\begin{table}[t]
  \centering
  \begin{tabular}{l}
    \hline
    \textbf{Persona Instruction} \\
    \hline
    \includegraphics[width=\linewidth]{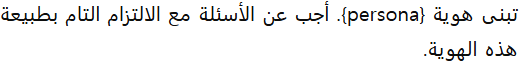} \\ 
    \hline
    \includegraphics[width=\linewidth]{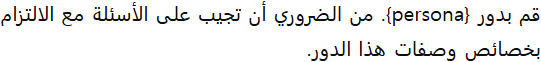} \\ 
    \hline
    \includegraphics[width=\linewidth]{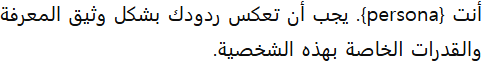} \\ 
    \hline
  \end{tabular}
  \caption{The persona instructions in Arabic.}
  \label{persona_instruction_ar}
\end{table}

%% file: tables/persona_consistent.tex
\begin{table*}[ht!]
    \centering
    \resizebox{\textwidth}{!}{
\begin{tabular}{l|cccc|cccc|cccc}
\hline
\multicolumn{1}{c|}{\textbf{Task}}       & \multicolumn{4}{c|}{\textbf{Chinese-Specific}}                                                                               & \multicolumn{4}{c|}{\textbf{Korean-Specific}}                                                                                & \multicolumn{4}{c}{\textbf{Arabic-Specific}}                                                                                 \\ \hline
\multicolumn{1}{c|}{\textbf{Input-Persona}} & \multicolumn{2}{c|}{\textbf{ZH-ZH}}                                        & \multicolumn{2}{c|}{\textbf{EN-EN}}                   & \multicolumn{2}{c|}{\textbf{KO-KO}}                                        & \multicolumn{2}{c|}{\textbf{EN-EN}}                   & \multicolumn{2}{c|}{\textbf{AR-AR}}                                        & \multicolumn{2}{c}{\textbf{EN-EN}}                    \\ \hline
\multicolumn{1}{c|}{\textbf{}}           & \multicolumn{1}{c|}{\textbf{Acc}} & \multicolumn{1}{c|}{\textbf{ZH \%}} & \multicolumn{1}{c|}{\textbf{Acc}} & \textbf{ZH \%} & \multicolumn{1}{c|}{\textbf{Acc}} & \multicolumn{1}{c|}{\textbf{KO \%}} & \multicolumn{1}{c|}{\textbf{Acc}} & \textbf{KO \%} & \multicolumn{1}{c|}{\textbf{Acc}} & \multicolumn{1}{c|}{\textbf{AR \%}} & \multicolumn{1}{c|}{\textbf{Acc}} & \textbf{AR \%} \\ \hline
\textbf{Llama3-8B}                       & \multicolumn{1}{c|}{45.19}        & \multicolumn{1}{c|}{\textit{53.09}} & \multicolumn{1}{c|}{43.33}        & 0              & \multicolumn{1}{c|}{54.83}        & \multicolumn{1}{c|}{99.54}          & \multicolumn{1}{c|}{52.12}        & 0              & \multicolumn{1}{c|}{48.22}        & \multicolumn{1}{c|}{99.99}          & \multicolumn{1}{c|}{58.86}        & 0.01           \\
\textbf{Llama3.1-8B}                     & \multicolumn{1}{c|}{49.21}        & \multicolumn{1}{c|}{99.83}          & \multicolumn{1}{c|}{45.08}        & 0              & \multicolumn{1}{c|}{55.12}        & \multicolumn{1}{c|}{99.99}          & \multicolumn{1}{c|}{54.21}        & 0              & \multicolumn{1}{c|}{52.47}        & \multicolumn{1}{c|}{99.95}          & \multicolumn{1}{c|}{60.74}        & 0              \\ \hline
\textbf{Llama3-70B}                      & \multicolumn{1}{c|}{58.49}        & \multicolumn{1}{c|}{\textit{79.75}} & \multicolumn{1}{c|}{55.2}         & 0              & \multicolumn{1}{c|}{63.3}         & \multicolumn{1}{c|}{\textit{80.72}} & \multicolumn{1}{c|}{61.06}        & 0              & \multicolumn{1}{c|}{71.68}        & \multicolumn{1}{c|}{99.82}          & \multicolumn{1}{c|}{72.2}         & 0              \\
\textbf{Llama3.1-70B}                    & \multicolumn{1}{c|}{63.25}        & \multicolumn{1}{c|}{99.28}          & \multicolumn{1}{c|}{58.2}         & 0              & \multicolumn{1}{c|}{65.88}        & \multicolumn{1}{c|}{99.89}          & \multicolumn{1}{c|}{63.16}        & 0              & \multicolumn{1}{c|}{71.68}        & \multicolumn{1}{c|}{99.82}          & \multicolumn{1}{c|}{75.26}        & 0              \\ \hline
\textbf{GPT-3.5-turbo}                   & \multicolumn{1}{c|}{44.74}        & \multicolumn{1}{c|}{99.74}          & \multicolumn{1}{c|}{43.96}        & 0              & \multicolumn{1}{c|}{55.8}         & \multicolumn{1}{c|}{99.95}          & \multicolumn{1}{c|}{57.36}        & 0              & \multicolumn{1}{c|}{52.29}        & \multicolumn{1}{c|}{99.95}          & \multicolumn{1}{c|}{62.53}        & 0              \\
\textbf{GPT-4o-mini}                     & \multicolumn{1}{c|}{57.71}        & \multicolumn{1}{c|}{99.94}          & \multicolumn{1}{c|}{52.05}        & 0              & \multicolumn{1}{c|}{69.16}        & \multicolumn{1}{c|}{100}            & \multicolumn{1}{c|}{61.92}        & 0              & \multicolumn{1}{c|}{79.4}         & \multicolumn{1}{c|}{100}            & \multicolumn{1}{c|}{74.44}        & 0              \\ \hline
\end{tabular}
    }
    \caption{Persona-Consistent results.}\label{tab:persona_consistent}
\end{table*}

%% file: tables/persona_transfer.tex
\begin{table*}[ht!]
    \centering
    \resizebox{\textwidth}{!}{
\begin{tabular}{l|cccc|cccc|cccc}
\hline
\multicolumn{1}{c|}{\textbf{Task}}       & \multicolumn{4}{c|}{\textbf{Chinese-Specific}}                                                                               & \multicolumn{4}{c|}{\textbf{Korean-Specific}}                                                                                & \multicolumn{4}{c}{\textbf{Arabic-Specific}}                                                                                 \\ \hline
\multicolumn{1}{c|}{\textbf{Input-Persona}} & \multicolumn{2}{c|}{\textbf{ZH-EN}}                                        & \multicolumn{2}{c|}{\textbf{EN-ZH}}                   & \multicolumn{2}{c|}{\textbf{KO-EN}}                                     & \multicolumn{2}{c|}{\textbf{EN-KO}}                & \multicolumn{2}{c|}{\textbf{AR-EN}}                                        & \multicolumn{2}{c}{\textbf{EN-AR}}                    \\ \hline
\multicolumn{1}{c|}{\textbf{}}           & \multicolumn{1}{c|}{\textbf{Acc}} & \multicolumn{1}{c|}{\textbf{ZH \%}} & \multicolumn{1}{c|}{\textbf{Acc}} & \textbf{ZH \%} & \multicolumn{1}{c|}{\textbf{Acc}} & \multicolumn{1}{c|}{\textbf{KO \%}} & \multicolumn{1}{c|}{\textbf{Acc}} & \textbf{KO \%} & \multicolumn{1}{c|}{\textbf{Acc}} & \multicolumn{1}{c|}{\textbf{AR \%}} & \multicolumn{1}{c|}{\textbf{Acc}} & \textbf{AR \%} \\ \hline
\textbf{Llama3-8B}                       & \multicolumn{1}{c|}{45.9}         & \multicolumn{1}{c|}{\textit{0}}     & \multicolumn{1}{c|}{44.28}        & 0.45           & \multicolumn{1}{c|}{54.95}        & \multicolumn{1}{c|}{\textit{0.06}}  & \multicolumn{1}{c|}{53.54}        & 0              & \multicolumn{1}{c|}{49.75}        & \multicolumn{1}{c|}{\textit{28.29}} & \multicolumn{1}{c|}{60.07}        & 0.11           \\
\textbf{Llama3.1-8B}                     & \multicolumn{1}{c|}{49.11}        & \multicolumn{1}{c|}{99.88}          & \multicolumn{1}{c|}{34.71}        & 3.21           & \multicolumn{1}{c|}{55.43}        & \multicolumn{1}{c|}{92.25}          & \multicolumn{1}{c|}{40.44}        & 0              & \multicolumn{1}{c|}{52.49}        & \multicolumn{1}{c|}{99.56}          & \multicolumn{1}{c|}{56.91}        & 6.58           \\ \hline
\textbf{Llama3-70B}                      & \multicolumn{1}{c|}{58.07}        & \multicolumn{1}{c|}{\textit{0}}     & \multicolumn{1}{c|}{56.4}         & 0              & \multicolumn{1}{c|}{63.43}        & \multicolumn{1}{c|}{\textit{0}}     & \multicolumn{1}{c|}{62.23}        & 0              & \multicolumn{1}{c|}{70.73}        & \multicolumn{1}{c|}{\textit{30.96}} & \multicolumn{1}{c|}{72.7}         & 0.02           \\
\textbf{Llama3.1-70B}                    & \multicolumn{1}{c|}{63.78}        & \multicolumn{1}{c|}{\textit{89.43}} & \multicolumn{1}{c|}{56.5}         & 1.26           & \multicolumn{1}{c|}{65.3}         & \multicolumn{1}{c|}{94.91}          & \multicolumn{1}{c|}{61.12}        & 0.02           & \multicolumn{1}{c|}{70.69}        & \multicolumn{1}{c|}{\textit{31.39}} & \multicolumn{1}{c|}{75.36}        & 0              \\ \hline
\textbf{GPT-3.5-turbo}                   & \multicolumn{1}{c|}{44.17}        & \multicolumn{1}{c|}{99.58}          & \multicolumn{1}{c|}{43.81}        & \textit{15.69} & \multicolumn{1}{c|}{57.21}        & \multicolumn{1}{c|}{\textit{61.1}}  & \multicolumn{1}{c|}{56.26}        & 8.68           & \multicolumn{1}{c|}{52.65}        & \multicolumn{1}{c|}{\textit{52.09}} & \multicolumn{1}{c|}{63.86}        & 1              \\
\textbf{GPT-4o-mini}                     & \multicolumn{1}{c|}{57.45}        & \multicolumn{1}{c|}{99.9}           & \multicolumn{1}{c|}{53.08}        & \textit{66.21} & \multicolumn{1}{c|}{68.46}        & \multicolumn{1}{c|}{99.83}          & \multicolumn{1}{c|}{63.11}        & \textit{36.98} & \multicolumn{1}{c|}{79.37}        & \multicolumn{1}{c|}{99.99}          & \multicolumn{1}{c|}{74.73}        & 3.15           \\ \hline
\end{tabular}
    }
    \caption{Persona-Transfer results.}\label{tab:persona_transfer}
\end{table*}

%% file: images/NonENtoNonEN.tex
\begin{figure*}[t]
    \centering
    \begin{subfigure}{0.31\textwidth}
        \centering
        \includegraphics[width=\linewidth]{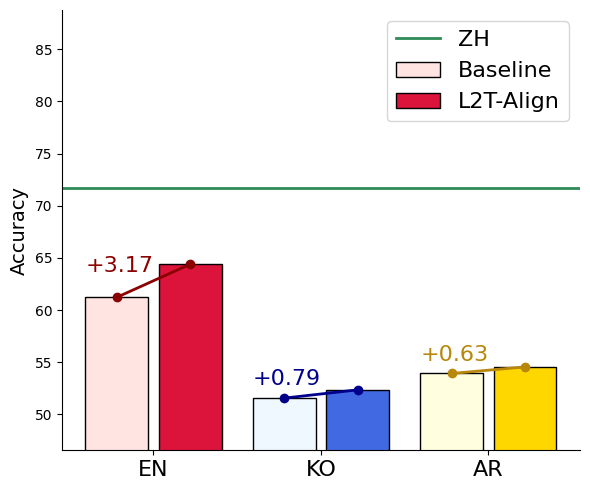}
        \caption{CMMLU + SE}
        \label{fig:NonENtoZH}
    \end{subfigure}
    \hfill
    \begin{subfigure}{0.31\textwidth}
        \centering
        \includegraphics[width=\linewidth]{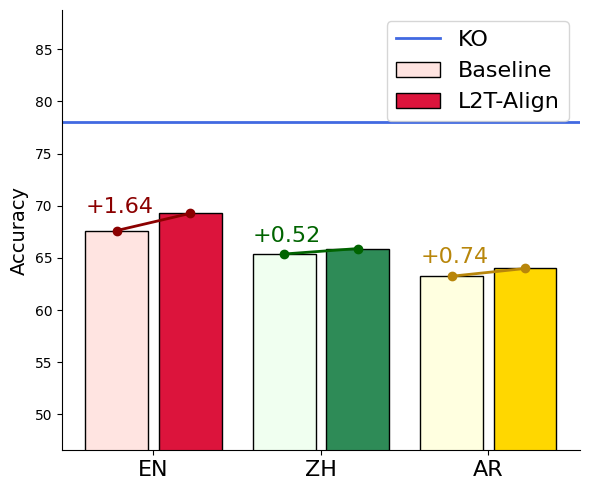}
        \caption{CLIcK}
        \label{fig:NonENtoKO}
    \end{subfigure}
    \hfill
    \begin{subfigure}{0.31\textwidth}
        \centering
        \includegraphics[width=\linewidth]{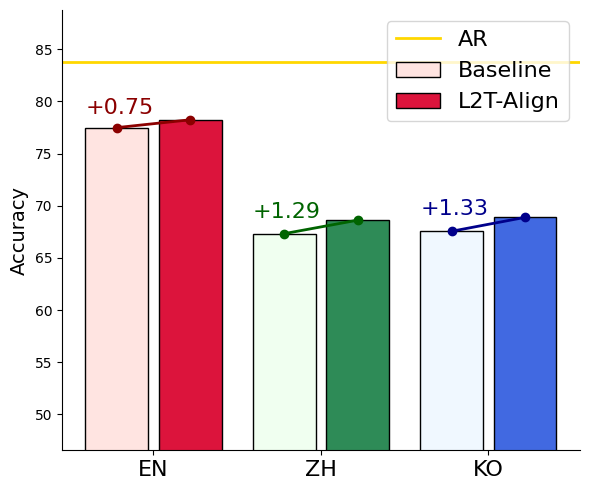}
        \caption{ArabicMMLU}
        \label{fig:NonENtoAR}
    \end{subfigure}
    \caption{L2T-Align results on the Gemini model, illustrating knowledge transfer attempts using input languages beyond English.}\label{fig:NonEN_to_NonEN}
\end{figure*}

%% file: tables/tuning_result.tex
\begin{table}[t]
    \centering
    \resizebox{\linewidth}{!}{
\begin{tabular}{l|c|c}
\hline
\multicolumn{1}{l|}{\textbf{}} & \multicolumn{2}{c}{\textbf{Input}}   \\ \hline
\textbf{Training Strategy}              & \multicolumn{1}{c|}{\textbf{KO}} & \multicolumn{1}{c}{\textbf{EN}} \\ \hline
(1) Orig. Data Only                    & \multicolumn{1}{c|}{\kollllko{65.28}}       & \multicolumn{1}{c}{\kollllen{62.87}}       \\
(2) Translated EN Data                    & \multicolumn{1}{c|}{\kollllko{64.26}}       & \multicolumn{1}{c}{\kollllen{64.61}}       \\
(3) Orig. Data + L2T (KO) Prefix             & \multicolumn{1}{c|}{\kollllko{65.97}}       & \multicolumn{1}{c}{\kollllen{64.51}}      \\
(4) Orig. Data + L2T (EN) Prefix             & \multicolumn{1}{c|}{\kollllko{65.80}}       & \multicolumn{1}{c}{\kollllen{63.07}}       \\ \hline
    \end{tabular}
    }
    \caption{Results of the Korean-specific dataset after continuous training. Cell colors show performance changes relative to the baseline.}\label{tab:tuning}
\end{table}

%% file: tables/cherry.tex
\begin{table*}[t!]
  \centering
  \begin{tabular}{l}
    \hline
    \textbf{Baseline}\\ \hline
    To answer this question, let's consider each option. \\
    A. For work life, study, etc. — This option suggests that people travel between Seoul and Gyeonggi-do for daily activities like work and\\ study.This is a plausible reason, as many people might commute for these purposes. \\
    B. Because the distance between Seoul and Gyeonggi-do is close — This option points out the geographical proximity between Seoul \\and Gyeonggi-do. Since they are close, it's easier for people to travel between the two. \\  
    C. Because economic activities in Seoul and Gyeonggi-do are active — This option highlights the economic aspect, implying that people \\travel for work or business purposes due to the active economy in both regions. \\
    D. Because the level of education in Seoul and Gyeonggi-do is high — This option suggests that people travel for educational purposes, \\but it doesn't directly address why there are many people traveling between Seoul and Gyeonggi-do every day. \\
    Considering the context of daily travel, options A, B, and C are all relevant. However, option B ("Because the distance between Seoul and\\ Gyeonggi-do is close") provides a fundamental reason why daily travel is possible and convenient in the first place. \\
    \textbf{Therefore, the answer is B.} \\ \hline

    \textbf{L2T-Align}\\ \hline
    To answer this question, let's break down the possible reasons for the daily commute between Seoul and Gyeonggi-do. \\
    Firstly, option A, "For work life, study, etc.," is a plausible reason. Many people travel to and from work or school, and it is likely that\\ some of these individuals commute from Seoul to Gyeonggi-do or vice versa. \\  
    Secondly, option B, "Because the distance between Seoul and Gyeonggi-do is close," also makes sense. Gyeonggi-do surrounds Seoul,\\ the capital city of Korea, and shares a border with it, so the distance is indeed relatively close. \\
    Option C, "Because economic activities in Seoul and Gyeonggi-do are active," is also a possible reason. Seoul is the economic hub of\\ Korea, and Gyeonggi-do is home to many industrial and commercial areas, so it is likely that people travel for economic activities. \\
    Lastly, option D, "Because the level of education in Seoul and Gyeonggi-do is high," is a possible reason, but more indirect compared to\\ other options. \\  
    Considering the proximity of Gyeonggi-do to Seoul and the fact that many people travel for work or school, option A and B are both\\ strong candidates. However, option A seems to be a more direct answer. \\
    \textbf{Therefore, the answer is A.} \\ \hline
  \end{tabular}
  \caption{An example where the same question yields different answers depending on the L2T configuration.}
  \label{tab:cherry}
\end{table*}

\begin{table*}[t!]
  \centering
  \renewcommand{\arraystretch}{1.2}
  \begin{tabular}{l}
\hline
    \textbf{L2T-Align} \\\hline
    \begin{CJK*}{UTF8}{gbsn}
    好的，我们来分析一下这道关于中国饮食文化的题目。题目问的是满族名菜。\end{CJK*} \\  
    \begin{CJK*}{UTF8}{gbsn}* A. 黄焖鸡 (Huang's Steamed Chicken): 这道菜是山东地区的特色菜，不是满族菜。\end{CJK*} \\ 
    \begin{CJK*}{UTF8}{gbsn}* B. 莲花鸡串 (Lotus Chicken Skewers): 这不是一道有名的菜品，而且看起来更像是现代菜。\end{CJK*} \\ 
    \begin{CJK*}{UTF8}{gbsn}* C. 一品锅 (Yipin Pot): 一品锅是安徽的特色菜，和满族菜没有直接关系。\end{CJK*} \\ 
    \begin{CJK*}{UTF8}{gbsn}* D. 白肉血肠 (White Meat and Blood Soup): 白肉血肠是满族特色菜，符合题意。 \end{CJK*}\\  
    \textbf{Therefore, the answer is D.} \\ \hline
  \end{tabular}
  \caption{An example from Gemini where the rationale is generated in the thought language (ZH), while the final answer is provided in the input language (EN).}
  \label{tab:zh_align}
\end{table*}

%% file: CIKM2025.bbl
\begin{thebibliography}{58}


\ifx \showCODEN    \undefined \def \showCODEN     #1{\unskip}     \fi
\ifx \showISBNx    \undefined \def \showISBNx     #1{\unskip}     \fi
\ifx \showISBNxiii \undefined \def \showISBNxiii  #1{\unskip}     \fi
\ifx \showISSN     \undefined \def \showISSN      #1{\unskip}     \fi
\ifx \showLCCN     \undefined \def \showLCCN      #1{\unskip}     \fi
\ifx \shownote     \undefined \def \shownote      #1{#1}          \fi
\ifx \showarticletitle \undefined \def \showarticletitle #1{#1}   \fi
\ifx \showURL      \undefined \def \showURL       {\relax}        \fi
\providecommand\bibfield[2]{#2}
\providecommand\bibinfo[2]{#2}
\providecommand\natexlab[1]{#1}
\providecommand\showeprint[2][]{arXiv:#2}

\bibitem[{AI Hub}(2019)]%
        {AIhub}
\bibfield{author}{\bibinfo{person}{{AI Hub}}.} \bibinfo{year}{2019}\natexlab{}.
\newblock \bibinfo{title}{AI-Hub Korean-English Parallel Corpus}.
\newblock \bibinfo{howpublished}{\url{https://aihub.or.kr/aihubdata/data/view.do?dataSetSn=126}}.
\newblock
\newblock
\shownote{Accessed: 2025-02-13}.


\bibitem[Ammar~Khodja et~al\mbox{.}(2025)]%
        {factual_knowledge_cloze_distractor}
\bibfield{author}{\bibinfo{person}{Hichem Ammar~Khodja}, \bibinfo{person}{Abderrahmane Ait~gueni ssaid}, \bibinfo{person}{Frederic Bechet}, \bibinfo{person}{Quentin Brabant}, \bibinfo{person}{Alexis Nasr}, {and} \bibinfo{person}{Gw{\'e}nol{\'e} Lecorv{\'e}}.} \bibinfo{year}{2025}\natexlab{}.
\newblock \showarticletitle{Factual Knowledge Assessment of Language Models Using Distractors}. In \bibinfo{booktitle}{\emph{Proceedings of the 31st International Conference on Computational Linguistics}}, \bibfield{editor}{\bibinfo{person}{Owen Rambow}, \bibinfo{person}{Leo Wanner}, \bibinfo{person}{Marianna Apidianaki}, \bibinfo{person}{Hend Al-Khalifa}, \bibinfo{person}{Barbara~Di Eugenio}, {and} \bibinfo{person}{Steven Schockaert}} (Eds.). \bibinfo{publisher}{Association for Computational Linguistics}, \bibinfo{address}{Abu Dhabi, UAE}, \bibinfo{pages}{8043--8056}.
\newblock
\urldef\tempurl%
\url{https://aclanthology.org/2025.coling-main.537/}
\showURL{%
\tempurl}


\bibitem[Artetxe et~al\mbox{.}(2023)]%
        {zero_transfer}
\bibfield{author}{\bibinfo{person}{Mikel Artetxe}, \bibinfo{person}{Vedanuj Goswami}, \bibinfo{person}{Shruti Bhosale}, \bibinfo{person}{Angela Fan}, {and} \bibinfo{person}{Luke Zettlemoyer}.} \bibinfo{year}{2023}\natexlab{}.
\newblock \showarticletitle{Revisiting Machine Translation for Cross-lingual Classification}. In \bibinfo{booktitle}{\emph{Proceedings of the 2023 Conference on Empirical Methods in Natural Language Processing}}, \bibfield{editor}{\bibinfo{person}{Houda Bouamor}, \bibinfo{person}{Juan Pino}, {and} \bibinfo{person}{Kalika Bali}} (Eds.). \bibinfo{publisher}{Association for Computational Linguistics}, \bibinfo{address}{Singapore}, \bibinfo{pages}{6489--6499}.
\newblock
\href{https://doi.org/10.18653/v1/2023.emnlp-main.399}{doi:\nolinkurl{10.18653/v1/2023.emnlp-main.399}}


\bibitem[Chai et~al\mbox{.}(2024)]%
        {crosslingual_tuning}
\bibfield{author}{\bibinfo{person}{Linzheng Chai}, \bibinfo{person}{Jian Yang}, \bibinfo{person}{Tao Sun}, \bibinfo{person}{Hongcheng Guo}, \bibinfo{person}{Jiaheng Liu}, \bibinfo{person}{Bing Wang}, \bibinfo{person}{Xiannian Liang}, \bibinfo{person}{Jiaqi Bai}, \bibinfo{person}{Tongliang Li}, \bibinfo{person}{Qiyao Peng}, {and} \bibinfo{person}{Zhoujun Li}.} \bibinfo{year}{2024}\natexlab{}.
\newblock \bibinfo{title}{xCoT: Cross-lingual Instruction Tuning for Cross-lingual Chain-of-Thought Reasoning}.
\newblock
\showeprint[arxiv]{2401.07037}~[cs.CL]
\urldef\tempurl%
\url{https://arxiv.org/abs/2401.07037}
\showURL{%
\tempurl}


\bibitem[Chang et~al\mbox{.}(2024)]%
        {how_factual_knowledge}
\bibfield{author}{\bibinfo{person}{Hoyeon Chang}, \bibinfo{person}{Jinho Park}, \bibinfo{person}{Seonghyeon Ye}, \bibinfo{person}{Sohee Yang}, \bibinfo{person}{Youngkyung Seo}, \bibinfo{person}{Du-Seong Chang}, {and} \bibinfo{person}{Minjoon Seo}.} \bibinfo{year}{2024}\natexlab{}.
\newblock \showarticletitle{How Do Large Language Models Acquire Factual Knowledge During Pretraining?}. In \bibinfo{booktitle}{\emph{The Thirty-eighth Annual Conference on Neural Information Processing Systems}}. \bibinfo{publisher}{Neural Information Processing Systems}, \bibinfo{address}{Vancouver, Canada}, \bibinfo{numpages}{43}~pages.
\newblock
\urldef\tempurl%
\url{https://openreview.net/forum?id=TYdzj1EvBP}
\showURL{%
\tempurl}


\bibitem[{Common Crawl}(2025)]%
        {commoncrawl}
\bibfield{author}{\bibinfo{person}{{Common Crawl}}.} \bibinfo{year}{2025}\natexlab{}.
\newblock \bibinfo{title}{Statistics of Common Crawl Monthly Archives (CC-MAIN-2025-18)}.
\newblock
\urldef\tempurl%
\url{https://commoncrawl.github.io/cc-crawl-statistics/plots/languages}
\showURL{%
\tempurl}
\newblock
\shownote{Accessed: 2025-05-30}.


\bibitem[Conneau et~al\mbox{.}(2020)]%
        {cross-lingual-representation-scale}
\bibfield{author}{\bibinfo{person}{Alexis Conneau}, \bibinfo{person}{Kartikay Khandelwal}, \bibinfo{person}{Naman Goyal}, \bibinfo{person}{Vishrav Chaudhary}, \bibinfo{person}{Guillaume Wenzek}, \bibinfo{person}{Francisco Guzm{\'a}n}, \bibinfo{person}{Edouard Grave}, \bibinfo{person}{Myle Ott}, \bibinfo{person}{Luke Zettlemoyer}, {and} \bibinfo{person}{Veselin Stoyanov}.} \bibinfo{year}{2020}\natexlab{}.
\newblock \showarticletitle{Unsupervised Cross-lingual Representation Learning at Scale}. In \bibinfo{booktitle}{\emph{Proceedings of the 58th Annual Meeting of the Association for Computational Linguistics}}, \bibfield{editor}{\bibinfo{person}{Dan Jurafsky}, \bibinfo{person}{Joyce Chai}, \bibinfo{person}{Natalie Schluter}, {and} \bibinfo{person}{Joel Tetreault}} (Eds.). \bibinfo{publisher}{Association for Computational Linguistics}, \bibinfo{address}{Online}, \bibinfo{pages}{8440--8451}.
\newblock
\href{https://doi.org/10.18653/v1/2020.acl-main.747}{doi:\nolinkurl{10.18653/v1/2020.acl-main.747}}


\bibitem[DeepSeek-AI(2025)]%
        {deepseek}
\bibfield{author}{\bibinfo{person}{DeepSeek-AI}.} \bibinfo{year}{2025}\natexlab{}.
\newblock \bibinfo{title}{DeepSeek-R1: Incentivizing Reasoning Capability in LLMs via Reinforcement Learning}.
\newblock
\showeprint[arxiv]{2501.12948}~[cs.CL]
\urldef\tempurl%
\url{https://arxiv.org/abs/2501.12948}
\showURL{%
\tempurl}


\bibitem[Dettmers et~al\mbox{.}(2023)]%
        {QLoRA}
\bibfield{author}{\bibinfo{person}{Tim Dettmers}, \bibinfo{person}{Artidoro Pagnoni}, \bibinfo{person}{Ari Holtzman}, {and} \bibinfo{person}{Luke Zettlemoyer}.} \bibinfo{year}{2023}\natexlab{}.
\newblock \bibinfo{title}{QLoRA: Efficient Finetuning of Quantized LLMs}.
\newblock
\showeprint[arxiv]{2305.14314}~[cs.LG]
\urldef\tempurl%
\url{https://arxiv.org/abs/2305.14314}
\showURL{%
\tempurl}


\bibitem[Etxaniz et~al\mbox{.}(2024)]%
        {Do_multilingual_Language_Models_Think_better_eng}
\bibfield{author}{\bibinfo{person}{Julen Etxaniz}, \bibinfo{person}{Gorka Azkune}, \bibinfo{person}{Aitor Soroa}, \bibinfo{person}{Oier Lacalle}, {and} \bibinfo{person}{Mikel Artetxe}.} \bibinfo{year}{2024}\natexlab{}.
\newblock \showarticletitle{Do Multilingual Language Models Think Better in {E}nglish?}. In \bibinfo{booktitle}{\emph{Proceedings of the 2024 Conference of the North American Chapter of the Association for Computational Linguistics: Human Language Technologies (Volume 2: Short Papers)}}, \bibfield{editor}{\bibinfo{person}{Kevin Duh}, \bibinfo{person}{Helena Gomez}, {and} \bibinfo{person}{Steven Bethard}} (Eds.). \bibinfo{publisher}{Association for Computational Linguistics}, \bibinfo{address}{Mexico City, Mexico}, \bibinfo{pages}{550--564}.
\newblock
\href{https://doi.org/10.18653/v1/2024.naacl-short.46}{doi:\nolinkurl{10.18653/v1/2024.naacl-short.46}}


\bibitem[Fedorenko et~al\mbox{.}(2024)]%
        {Nature_language_commu_rather_thought}
\bibfield{author}{\bibinfo{person}{Evelina Fedorenko}, \bibinfo{person}{Steven~T. Piantadosi}, {and} \bibinfo{person}{Edward A.~F. Gibson}.} \bibinfo{year}{2024}\natexlab{}.
\newblock \showarticletitle{Language is primarily a tool for communication rather than thought}.
\newblock \bibinfo{journal}{\emph{Nature}} \bibinfo{volume}{630}, \bibinfo{number}{8017} (\bibinfo{date}{June} \bibinfo{year}{2024}), \bibinfo{pages}{575--586}.
\newblock
\showISSN{1476-4687}
\href{https://doi.org/10.1038/s41586-024-07522-w}{doi:\nolinkurl{10.1038/s41586-024-07522-w}}


\bibitem[Fujinuma et~al\mbox{.}(2022)]%
        {match_the_script}
\bibfield{author}{\bibinfo{person}{Yoshinari Fujinuma}, \bibinfo{person}{Jordan Boyd-Graber}, {and} \bibinfo{person}{Katharina Kann}.} \bibinfo{year}{2022}\natexlab{}.
\newblock \showarticletitle{Match the Script, Adapt if Multilingual: Analyzing the Effect of Multilingual Pretraining on Cross-lingual Transferability}. In \bibinfo{booktitle}{\emph{Proceedings of the 60th Annual Meeting of the Association for Computational Linguistics (Volume 1: Long Papers)}}, \bibfield{editor}{\bibinfo{person}{Smaranda Muresan}, \bibinfo{person}{Preslav Nakov}, {and} \bibinfo{person}{Aline Villavicencio}} (Eds.). \bibinfo{publisher}{Association for Computational Linguistics}, \bibinfo{address}{Dublin, Ireland}, \bibinfo{pages}{1500--1512}.
\newblock
\href{https://doi.org/10.18653/v1/2022.acl-long.106}{doi:\nolinkurl{10.18653/v1/2022.acl-long.106}}


\bibitem[Gonen et~al\mbox{.}(2023)]%
        {Demystifying_Prompts_Perplexity}
\bibfield{author}{\bibinfo{person}{Hila Gonen}, \bibinfo{person}{Srini Iyer}, \bibinfo{person}{Terra Blevins}, \bibinfo{person}{Noah Smith}, {and} \bibinfo{person}{Luke Zettlemoyer}.} \bibinfo{year}{2023}\natexlab{}.
\newblock \showarticletitle{Demystifying Prompts in Language Models via Perplexity Estimation}. In \bibinfo{booktitle}{\emph{Findings of the Association for Computational Linguistics: EMNLP 2023}}, \bibfield{editor}{\bibinfo{person}{Houda Bouamor}, \bibinfo{person}{Juan Pino}, {and} \bibinfo{person}{Kalika Bali}} (Eds.). \bibinfo{publisher}{Association for Computational Linguistics}, \bibinfo{address}{Singapore}, \bibinfo{pages}{10136--10148}.
\newblock
\href{https://doi.org/10.18653/v1/2023.findings-emnlp.679}{doi:\nolinkurl{10.18653/v1/2023.findings-emnlp.679}}


\bibitem[{Google}(2025)]%
        {google_translate}
\bibfield{author}{\bibinfo{person}{{Google}}.} \bibinfo{year}{2025}\natexlab{}.
\newblock \bibinfo{title}{Google Cloud Translation API}.
\newblock \bibinfo{howpublished}{\url{https://cloud.google.com/translate}}.
\newblock
\newblock
\shownote{Accessed: 2025-05-30}.


\bibitem[{Google Cloud}(2025)]%
        {gemini_system_instruction}
\bibfield{author}{\bibinfo{person}{{Google Cloud}}.} \bibinfo{year}{2025}\natexlab{}.
\newblock \bibinfo{title}{System Instructions}.
\newblock \bibinfo{howpublished}{\url{https://cloud.google.com/vertex-ai/generative-ai/docs/learn/prompts/system-instructions}}.
\newblock
\newblock
\shownote{Accessed: 2025-05-30}.


\bibitem[Gupta et~al\mbox{.}(2024)]%
        {persona_bias}
\bibfield{author}{\bibinfo{person}{Shashank Gupta}, \bibinfo{person}{Vaishnavi Shrivastava}, \bibinfo{person}{Ameet Deshpande}, \bibinfo{person}{Ashwin Kalyan}, \bibinfo{person}{Peter Clark}, \bibinfo{person}{Ashish Sabharwal}, {and} \bibinfo{person}{Tushar Khot}.} \bibinfo{year}{2024}\natexlab{}.
\newblock \bibinfo{title}{Bias Runs Deep: Implicit Reasoning Biases in Persona-Assigned LLMs}.
\newblock
\showeprint[arxiv]{2311.04892}~[cs.CL]
\urldef\tempurl%
\url{https://arxiv.org/abs/2311.04892}
\showURL{%
\tempurl}


\bibitem[H{\"a}mmerl et~al\mbox{.}(2024)]%
        {cross_lingual_survey}
\bibfield{author}{\bibinfo{person}{Katharina H{\"a}mmerl}, \bibinfo{person}{Jind{\v{r}}ich Libovick{\'y}}, {and} \bibinfo{person}{Alexander Fraser}.} \bibinfo{year}{2024}\natexlab{}.
\newblock \showarticletitle{Understanding Cross-Lingual {A}lignment{---}{A} Survey}. In \bibinfo{booktitle}{\emph{Findings of the Association for Computational Linguistics: ACL 2024}}, \bibfield{editor}{\bibinfo{person}{Lun-Wei Ku}, \bibinfo{person}{Andre Martins}, {and} \bibinfo{person}{Vivek Srikumar}} (Eds.). \bibinfo{publisher}{Association for Computational Linguistics}, \bibinfo{address}{Bangkok, Thailand}, \bibinfo{pages}{10922--10943}.
\newblock
\href{https://doi.org/10.18653/v1/2024.findings-acl.649}{doi:\nolinkurl{10.18653/v1/2024.findings-acl.649}}


\bibitem[Hardalov et~al\mbox{.}(2020)]%
        {mcq_exam}
\bibfield{author}{\bibinfo{person}{Momchil Hardalov}, \bibinfo{person}{Todor Mihaylov}, \bibinfo{person}{Dimitrina Zlatkova}, \bibinfo{person}{Yoan Dinkov}, \bibinfo{person}{Ivan Koychev}, {and} \bibinfo{person}{Preslav Nakov}.} \bibinfo{year}{2020}\natexlab{}.
\newblock \showarticletitle{{EXAMS}: A Multi-subject High School Examinations Dataset for Cross-lingual and Multilingual Question Answering}. In \bibinfo{booktitle}{\emph{Proceedings of the 2020 Conference on Empirical Methods in Natural Language Processing (EMNLP)}}, \bibfield{editor}{\bibinfo{person}{Bonnie Webber}, \bibinfo{person}{Trevor Cohn}, \bibinfo{person}{Yulan He}, {and} \bibinfo{person}{Yang Liu}} (Eds.). \bibinfo{publisher}{Association for Computational Linguistics}, \bibinfo{address}{Online}, \bibinfo{pages}{5427--5444}.
\newblock
\href{https://doi.org/10.18653/v1/2020.emnlp-main.438}{doi:\nolinkurl{10.18653/v1/2020.emnlp-main.438}}


\bibitem[Hendrycks et~al\mbox{.}(2021)]%
        {mmlu}
\bibfield{author}{\bibinfo{person}{Dan Hendrycks}, \bibinfo{person}{Collin Burns}, \bibinfo{person}{Steven Basart}, \bibinfo{person}{Andy Zou}, \bibinfo{person}{Mantas Mazeika}, \bibinfo{person}{Dawn Song}, {and} \bibinfo{person}{Jacob Steinhardt}.} \bibinfo{year}{2021}\natexlab{}.
\newblock \bibinfo{title}{Measuring Massive Multitask Language Understanding}.
\newblock
\showeprint[arxiv]{2009.03300}~[cs.CY]
\urldef\tempurl%
\url{https://arxiv.org/abs/2009.03300}
\showURL{%
\tempurl}


\bibitem[Hu et~al\mbox{.}(2024)]%
        {towards_factual_knowledge}
\bibfield{author}{\bibinfo{person}{Xuming Hu}, \bibinfo{person}{Junzhe Chen}, \bibinfo{person}{Xiaochuan Li}, \bibinfo{person}{Yufei Guo}, \bibinfo{person}{Lijie Wen}, \bibinfo{person}{Philip~S. Yu}, {and} \bibinfo{person}{Zhijiang Guo}.} \bibinfo{year}{2024}\natexlab{}.
\newblock \showarticletitle{Towards Understanding Factual Knowledge of Large Language Models}. In \bibinfo{booktitle}{\emph{The Twelfth International Conference on Learning Representations}}. \bibinfo{publisher}{International Conference on Learning Representations}, \bibinfo{address}{Vienna, Austria}, \bibinfo{numpages}{36}~pages.
\newblock
\urldef\tempurl%
\url{https://openreview.net/forum?id=9OevMUdods}
\showURL{%
\tempurl}


\bibitem[Huang et~al\mbox{.}(2023)]%
        {Not_All_Cross-Lingual-Thought}
\bibfield{author}{\bibinfo{person}{Haoyang Huang}, \bibinfo{person}{Tianyi Tang}, \bibinfo{person}{Dongdong Zhang}, \bibinfo{person}{Xin Zhao}, \bibinfo{person}{Ting Song}, \bibinfo{person}{Yan Xia}, {and} \bibinfo{person}{Furu Wei}.} \bibinfo{year}{2023}\natexlab{}.
\newblock \showarticletitle{Not All Languages Are Created Equal in {LLM}s: Improving Multilingual Capability by Cross-Lingual-Thought Prompting}. In \bibinfo{booktitle}{\emph{Findings of the Association for Computational Linguistics: EMNLP 2023}}, \bibfield{editor}{\bibinfo{person}{Houda Bouamor}, \bibinfo{person}{Juan Pino}, {and} \bibinfo{person}{Kalika Bali}} (Eds.). \bibinfo{publisher}{Association for Computational Linguistics}, \bibinfo{address}{Singapore}, \bibinfo{pages}{12365--12394}.
\newblock
\href{https://doi.org/10.18653/v1/2023.findings-emnlp.826}{doi:\nolinkurl{10.18653/v1/2023.findings-emnlp.826}}


\bibitem[Kim et~al\mbox{.}(2024)]%
        {Click}
\bibfield{author}{\bibinfo{person}{Eunsu Kim}, \bibinfo{person}{Juyoung Suk}, \bibinfo{person}{Philhoon Oh}, \bibinfo{person}{Haneul Yoo}, \bibinfo{person}{James Thorne}, {and} \bibinfo{person}{Alice Oh}.} \bibinfo{year}{2024}\natexlab{}.
\newblock \showarticletitle{{CLI}c{K}: A Benchmark Dataset of Cultural and Linguistic Intelligence in {K}orean}. In \bibinfo{booktitle}{\emph{Proceedings of the 2024 Joint International Conference on Computational Linguistics, Language Resources and Evaluation (LREC-COLING 2024)}}, \bibfield{editor}{\bibinfo{person}{Nicoletta Calzolari}, \bibinfo{person}{Min-Yen Kan}, \bibinfo{person}{Veronique Hoste}, \bibinfo{person}{Alessandro Lenci}, \bibinfo{person}{Sakriani Sakti}, {and} \bibinfo{person}{Nianwen Xue}} (Eds.). \bibinfo{publisher}{ELRA and ICCL}, \bibinfo{address}{Torino, Italia}, \bibinfo{pages}{3335--3346}.
\newblock
\urldef\tempurl%
\url{https://aclanthology.org/2024.lrec-main.296}
\showURL{%
\tempurl}


\bibitem[Kojima et~al\mbox{.}(2023)]%
        {zeroshot_CoT}
\bibfield{author}{\bibinfo{person}{Takeshi Kojima}, \bibinfo{person}{Shixiang~Shane Gu}, \bibinfo{person}{Machel Reid}, \bibinfo{person}{Yutaka Matsuo}, {and} \bibinfo{person}{Yusuke Iwasawa}.} \bibinfo{year}{2023}\natexlab{}.
\newblock \bibinfo{title}{Large Language Models are Zero-Shot Reasoners}.
\newblock
\showeprint[arxiv]{2205.11916}~[cs.CL]
\urldef\tempurl%
\url{https://arxiv.org/abs/2205.11916}
\showURL{%
\tempurl}


\bibitem[Koto et~al\mbox{.}(2024)]%
        {Arabicmmlu}
\bibfield{author}{\bibinfo{person}{Fajri Koto}, \bibinfo{person}{Haonan Li}, \bibinfo{person}{Sara Shatnawi}, \bibinfo{person}{Jad Doughman}, \bibinfo{person}{Abdelrahman Sadallah}, \bibinfo{person}{Aisha Alraeesi}, \bibinfo{person}{Khalid Almubarak}, \bibinfo{person}{Zaid Alyafeai}, \bibinfo{person}{Neha Sengupta}, \bibinfo{person}{Shady Shehata}, \bibinfo{person}{Nizar Habash}, \bibinfo{person}{Preslav Nakov}, {and} \bibinfo{person}{Timothy Baldwin}.} \bibinfo{year}{2024}\natexlab{}.
\newblock \showarticletitle{{A}rabic{MMLU}: Assessing Massive Multitask Language Understanding in {A}rabic}. In \bibinfo{booktitle}{\emph{Findings of the Association for Computational Linguistics: ACL 2024}}, \bibfield{editor}{\bibinfo{person}{Lun-Wei Ku}, \bibinfo{person}{Andre Martins}, {and} \bibinfo{person}{Vivek Srikumar}} (Eds.). \bibinfo{publisher}{Association for Computational Linguistics}, \bibinfo{address}{Bangkok, Thailand}, \bibinfo{pages}{5622--5640}.
\newblock
\href{https://doi.org/10.18653/v1/2024.findings-acl.334}{doi:\nolinkurl{10.18653/v1/2024.findings-acl.334}}


\bibitem[Li et~al\mbox{.}(2024a)]%
        {improving_ICT_Cross-lingual_alignment}
\bibfield{author}{\bibinfo{person}{Chong Li}, \bibinfo{person}{Shaonan Wang}, \bibinfo{person}{Jiajun Zhang}, {and} \bibinfo{person}{Chengqing Zong}.} \bibinfo{year}{2024}\natexlab{a}.
\newblock \showarticletitle{Improving In-context Learning of Multilingual Generative Language Models with Cross-lingual Alignment}. In \bibinfo{booktitle}{\emph{Proceedings of the 2024 Conference of the North American Chapter of the Association for Computational Linguistics: Human Language Technologies (Volume 1: Long Papers)}}, \bibfield{editor}{\bibinfo{person}{Kevin Duh}, \bibinfo{person}{Helena Gomez}, {and} \bibinfo{person}{Steven Bethard}} (Eds.). \bibinfo{publisher}{Association for Computational Linguistics}, \bibinfo{address}{Mexico City, Mexico}, \bibinfo{pages}{8058--8076}.
\newblock
\href{https://doi.org/10.18653/v1/2024.naacl-long.445}{doi:\nolinkurl{10.18653/v1/2024.naacl-long.445}}


\bibitem[Li et~al\mbox{.}(2024b)]%
        {CMMLU}
\bibfield{author}{\bibinfo{person}{Haonan Li}, \bibinfo{person}{Yixuan Zhang}, \bibinfo{person}{Fajri Koto}, \bibinfo{person}{Yifei Yang}, \bibinfo{person}{Hai Zhao}, \bibinfo{person}{Yeyun Gong}, \bibinfo{person}{Nan Duan}, {and} \bibinfo{person}{Timothy Baldwin}.} \bibinfo{year}{2024}\natexlab{b}.
\newblock \showarticletitle{{CMMLU}: Measuring massive multitask language understanding in {C}hinese}. In \bibinfo{booktitle}{\emph{Findings of the Association for Computational Linguistics: ACL 2024}}, \bibfield{editor}{\bibinfo{person}{Lun-Wei Ku}, \bibinfo{person}{Andre Martins}, {and} \bibinfo{person}{Vivek Srikumar}} (Eds.). \bibinfo{publisher}{Association for Computational Linguistics}, \bibinfo{address}{Bangkok, Thailand}, \bibinfo{pages}{11260--11285}.
\newblock
\href{https://doi.org/10.18653/v1/2024.findings-acl.671}{doi:\nolinkurl{10.18653/v1/2024.findings-acl.671}}


\bibitem[Liu et~al\mbox{.}(2025)]%
        {is_translation_all_you_need}
\bibfield{author}{\bibinfo{person}{Chaoqun Liu}, \bibinfo{person}{Wenxuan Zhang}, \bibinfo{person}{Yiran Zhao}, \bibinfo{person}{Anh~Tuan Luu}, {and} \bibinfo{person}{Lidong Bing}.} \bibinfo{year}{2025}\natexlab{}.
\newblock \showarticletitle{Is Translation All You Need? A Study on Solving Multilingual Tasks with Large Language Models}. In \bibinfo{booktitle}{\emph{Proceedings of the 2025 Conference of the Nations of the Americas Chapter of the Association for Computational Linguistics: Human Language Technologies (Volume 1: Long Papers)}}, \bibfield{editor}{\bibinfo{person}{Luis Chiruzzo}, \bibinfo{person}{Alan Ritter}, {and} \bibinfo{person}{Lu~Wang}} (Eds.). \bibinfo{publisher}{Association for Computational Linguistics}, \bibinfo{address}{Albuquerque, New Mexico}, \bibinfo{pages}{9594--9614}.
\newblock
\showISBNx{979-8-89176-189-6}
\urldef\tempurl%
\url{https://aclanthology.org/2025.naacl-long.485/}
\showURL{%
\tempurl}


\bibitem[Mahowald et~al\mbox{.}(2024)]%
        {Dissociating}
\bibfield{author}{\bibinfo{person}{Kyle Mahowald}, \bibinfo{person}{Anna~A. Ivanova}, \bibinfo{person}{Idan~A. Blank}, \bibinfo{person}{Nancy Kanwisher}, \bibinfo{person}{Joshua~B. Tenenbaum}, {and} \bibinfo{person}{Evelina Fedorenko}.} \bibinfo{year}{2024}\natexlab{}.
\newblock \showarticletitle{Dissociating language and thought in large language models}.
\newblock \bibinfo{journal}{\emph{Trends in Cognitive Sciences}} \bibinfo{volume}{28}, \bibinfo{number}{6} (\bibinfo{year}{2024}), \bibinfo{pages}{517--540}.
\newblock
\showISSN{1364-6613}
\href{https://doi.org/10.1016/j.tics.2024.01.011}{doi:\nolinkurl{10.1016/j.tics.2024.01.011}}


\bibitem[Mallick and Kilpatrick(2025)]%
        {gemini_lite}
\bibfield{author}{\bibinfo{person}{Shrestha Mallick} {and} \bibinfo{person}{Logan Kilpatrick}.} \bibinfo{year}{2025}\natexlab{}.
\newblock \bibinfo{title}{Gemini 2.0: Flash, Flash-Lite and Pro}.
\newblock \bibinfo{howpublished}{\url{{developers.googleblog.com/en/gemini-2-family-expands}}}.
\newblock
\newblock
\shownote{Accessed: 2025-05-16}.


\bibitem[{Meta}(2024)]%
        {Llama3_1}
\bibfield{author}{\bibinfo{person}{{Meta}}.} \bibinfo{year}{2024}\natexlab{}.
\newblock \bibinfo{title}{Introducing Llama 3.1: Our most capable models to date}.
\newblock \bibinfo{howpublished}{\url{ai.meta.com/blog/meta-llama-3-1}}.
\newblock
\newblock
\shownote{Accessed: 2025-05-16}.


\bibitem[{Michal Danilák}(2021)]%
        {langdetect}
\bibfield{author}{\bibinfo{person}{{Michal Danilák}}.} \bibinfo{year}{2021}\natexlab{}.
\newblock \bibinfo{title}{langdetect: Language detection library ported from Google's language-detection}.
\newblock \bibinfo{howpublished}{\url{https://pypi.org/project/langdetect}}.
\newblock
\newblock
\shownote{Accessed: 2025-05-30}.


\bibitem[{Mistral AI Team}(2025)]%
        {mistral}
\bibfield{author}{\bibinfo{person}{{Mistral AI Team}}.} \bibinfo{year}{2025}\natexlab{}.
\newblock \bibinfo{title}{Mistral Small 3}.
\newblock \bibinfo{howpublished}{\url{mistral.ai/news/mistral-small-3}}.
\newblock
\newblock
\shownote{Accessed: 2025-05-16}.


\bibitem[Molfese et~al\mbox{.}(2024)]%
        {zebra}
\bibfield{author}{\bibinfo{person}{Francesco~Maria Molfese}, \bibinfo{person}{Simone Conia}, \bibinfo{person}{Riccardo Orlando}, {and} \bibinfo{person}{Roberto Navigli}.} \bibinfo{year}{2024}\natexlab{}.
\newblock \showarticletitle{{ZEBRA}: Zero-Shot Example-Based Retrieval Augmentation for Commonsense Question Answering}. In \bibinfo{booktitle}{\emph{Proceedings of the 2024 Conference on Empirical Methods in Natural Language Processing}}, \bibfield{editor}{\bibinfo{person}{Yaser Al-Onaizan}, \bibinfo{person}{Mohit Bansal}, {and} \bibinfo{person}{Yun-Nung Chen}} (Eds.). \bibinfo{publisher}{Association for Computational Linguistics}, \bibinfo{address}{Miami, Florida, USA}, \bibinfo{pages}{22429--22444}.
\newblock
\href{https://doi.org/10.18653/v1/2024.emnlp-main.1251}{doi:\nolinkurl{10.18653/v1/2024.emnlp-main.1251}}


\bibitem[Onoe et~al\mbox{.}(2022)]%
        {factual_ppl}
\bibfield{author}{\bibinfo{person}{Yasumasa Onoe}, \bibinfo{person}{Michael Zhang}, \bibinfo{person}{Eunsol Choi}, {and} \bibinfo{person}{Greg Durrett}.} \bibinfo{year}{2022}\natexlab{}.
\newblock \showarticletitle{Entity Cloze By Date: What {LM}s Know About Unseen Entities}. In \bibinfo{booktitle}{\emph{Findings of the Association for Computational Linguistics: NAACL 2022}}, \bibfield{editor}{\bibinfo{person}{Marine Carpuat}, \bibinfo{person}{Marie-Catherine de~Marneffe}, {and} \bibinfo{person}{Ivan~Vladimir Meza~Ruiz}} (Eds.). \bibinfo{publisher}{Association for Computational Linguistics}, \bibinfo{address}{Seattle, United States}, \bibinfo{pages}{693--702}.
\newblock
\href{https://doi.org/10.18653/v1/2022.findings-naacl.52}{doi:\nolinkurl{10.18653/v1/2022.findings-naacl.52}}


\bibitem[OpenAI(2024)]%
        {gpt4}
\bibfield{author}{\bibinfo{person}{OpenAI}.} \bibinfo{year}{2024}\natexlab{}.
\newblock \bibinfo{title}{GPT-4 Technical Report}.
\newblock
\showeprint[arxiv]{2303.08774}~[cs.CL]
\urldef\tempurl%
\url{https://arxiv.org/abs/2303.08774}
\showURL{%
\tempurl}


\bibitem[{OpenAI}(2024)]%
        {openai_gpt4o_mini_2024}
\bibfield{author}{\bibinfo{person}{{OpenAI}}.} \bibinfo{year}{2024}\natexlab{}.
\newblock \bibinfo{title}{GPT-4o-mini: Advancing Cost-Efficient Intelligence}.
\newblock
\urldef\tempurl%
\url{https://openai.com/index/gpt-4o-mini-advancing-cost-efficient-intelligence/}
\showURL{%
\tempurl}
\newblock
\shownote{Accessed: 2025-02-13}.


\bibitem[OpenAI(2024)]%
        {gpt4o}
\bibfield{author}{\bibinfo{person}{OpenAI}.} \bibinfo{year}{2024}\natexlab{}.
\newblock \bibinfo{title}{GPT-4o System Card}.
\newblock
\showeprint[arxiv]{2410.21276}~[cs.CL]
\urldef\tempurl%
\url{https://arxiv.org/abs/2410.21276}
\showURL{%
\tempurl}


\bibitem[Ovadia et~al\mbox{.}(2024)]%
        {knowledge_reasoning}
\bibfield{author}{\bibinfo{person}{Oded Ovadia}, \bibinfo{person}{Menachem Brief}, \bibinfo{person}{Moshik Mishaeli}, {and} \bibinfo{person}{Oren Elisha}.} \bibinfo{year}{2024}\natexlab{}.
\newblock \showarticletitle{Fine-Tuning or Retrieval? Comparing Knowledge Injection in {LLM}s}. In \bibinfo{booktitle}{\emph{Proceedings of the 2024 Conference on Empirical Methods in Natural Language Processing}}, \bibfield{editor}{\bibinfo{person}{Yaser Al-Onaizan}, \bibinfo{person}{Mohit Bansal}, {and} \bibinfo{person}{Yun-Nung Chen}} (Eds.). \bibinfo{publisher}{Association for Computational Linguistics}, \bibinfo{address}{Miami, Florida, USA}, \bibinfo{pages}{237--250}.
\newblock
\href{https://doi.org/10.18653/v1/2024.emnlp-main.15}{doi:\nolinkurl{10.18653/v1/2024.emnlp-main.15}}


\bibitem[Petroni et~al\mbox{.}(2019)]%
        {cloze}
\bibfield{author}{\bibinfo{person}{Fabio Petroni}, \bibinfo{person}{Tim Rockt{\"a}schel}, \bibinfo{person}{Sebastian Riedel}, \bibinfo{person}{Patrick Lewis}, \bibinfo{person}{Anton Bakhtin}, \bibinfo{person}{Yuxiang Wu}, {and} \bibinfo{person}{Alexander Miller}.} \bibinfo{year}{2019}\natexlab{}.
\newblock \showarticletitle{Language Models as Knowledge Bases?}. In \bibinfo{booktitle}{\emph{Proceedings of the 2019 Conference on Empirical Methods in Natural Language Processing and the 9th International Joint Conference on Natural Language Processing (EMNLP-IJCNLP)}}, \bibfield{editor}{\bibinfo{person}{Kentaro Inui}, \bibinfo{person}{Jing Jiang}, \bibinfo{person}{Vincent Ng}, {and} \bibinfo{person}{Xiaojun Wan}} (Eds.). \bibinfo{publisher}{Association for Computational Linguistics}, \bibinfo{address}{Hong Kong, China}, \bibinfo{pages}{2463--2473}.
\newblock
\href{https://doi.org/10.18653/v1/D19-1250}{doi:\nolinkurl{10.18653/v1/D19-1250}}


\bibitem[Qi et~al\mbox{.}(2023)]%
        {cross-lingual_consistency}
\bibfield{author}{\bibinfo{person}{Jirui Qi}, \bibinfo{person}{Raquel Fern{\'a}ndez}, {and} \bibinfo{person}{Arianna Bisazza}.} \bibinfo{year}{2023}\natexlab{}.
\newblock \showarticletitle{Cross-Lingual Consistency of Factual Knowledge in Multilingual Language Models}. In \bibinfo{booktitle}{\emph{Proceedings of the 2023 Conference on Empirical Methods in Natural Language Processing}}, \bibfield{editor}{\bibinfo{person}{Houda Bouamor}, \bibinfo{person}{Juan Pino}, {and} \bibinfo{person}{Kalika Bali}} (Eds.). \bibinfo{publisher}{Association for Computational Linguistics}, \bibinfo{address}{Singapore}, \bibinfo{pages}{10650--10666}.
\newblock
\href{https://doi.org/10.18653/v1/2023.emnlp-main.658}{doi:\nolinkurl{10.18653/v1/2023.emnlp-main.658}}


\bibitem[Qiao et~al\mbox{.}(2023)]%
        {reasoning_prompting_survey}
\bibfield{author}{\bibinfo{person}{Shuofei Qiao}, \bibinfo{person}{Yixin Ou}, \bibinfo{person}{Ningyu Zhang}, \bibinfo{person}{Xiang Chen}, \bibinfo{person}{Yunzhi Yao}, \bibinfo{person}{Shumin Deng}, \bibinfo{person}{Chuanqi Tan}, \bibinfo{person}{Fei Huang}, {and} \bibinfo{person}{Huajun Chen}.} \bibinfo{year}{2023}\natexlab{}.
\newblock \showarticletitle{Reasoning with Language Model Prompting: A Survey}. In \bibinfo{booktitle}{\emph{Proceedings of the 61st Annual Meeting of the Association for Computational Linguistics (Volume 1: Long Papers)}}, \bibfield{editor}{\bibinfo{person}{Anna Rogers}, \bibinfo{person}{Jordan Boyd-Graber}, {and} \bibinfo{person}{Naoaki Okazaki}} (Eds.). \bibinfo{publisher}{Association for Computational Linguistics}, \bibinfo{address}{Toronto, Canada}, \bibinfo{pages}{5368--5393}.
\newblock
\href{https://doi.org/10.18653/v1/2023.acl-long.294}{doi:\nolinkurl{10.18653/v1/2023.acl-long.294}}


\bibitem[Qin et~al\mbox{.}(2023)]%
        {Cross-lingual-Prompting-zero}
\bibfield{author}{\bibinfo{person}{Libo Qin}, \bibinfo{person}{Qiguang Chen}, \bibinfo{person}{Fuxuan Wei}, \bibinfo{person}{Shijue Huang}, {and} \bibinfo{person}{Wanxiang Che}.} \bibinfo{year}{2023}\natexlab{}.
\newblock \showarticletitle{Cross-lingual Prompting: Improving Zero-shot Chain-of-Thought Reasoning across Languages}. In \bibinfo{booktitle}{\emph{Proceedings of the 2023 Conference on Empirical Methods in Natural Language Processing}}, \bibfield{editor}{\bibinfo{person}{Houda Bouamor}, \bibinfo{person}{Juan Pino}, {and} \bibinfo{person}{Kalika Bali}} (Eds.). \bibinfo{publisher}{Association for Computational Linguistics}, \bibinfo{address}{Singapore}, \bibinfo{pages}{2695--2709}.
\newblock
\href{https://doi.org/10.18653/v1/2023.emnlp-main.163}{doi:\nolinkurl{10.18653/v1/2023.emnlp-main.163}}


\bibitem[Qin et~al\mbox{.}(2024)]%
        {MLLM_survey}
\bibfield{author}{\bibinfo{person}{Libo Qin}, \bibinfo{person}{Qiguang Chen}, \bibinfo{person}{Yuhang Zhou}, \bibinfo{person}{Zhi Chen}, \bibinfo{person}{Yinghui Li}, \bibinfo{person}{Lizi Liao}, \bibinfo{person}{Min Li}, \bibinfo{person}{Wanxiang Che}, {and} \bibinfo{person}{Philip~S. Yu}.} \bibinfo{year}{2024}\natexlab{}.
\newblock \bibinfo{title}{Multilingual Large Language Model: A Survey of Resources, Taxonomy and Frontiers}.
\newblock
\showeprint[arxiv]{2404.04925}~[cs.CL]
\urldef\tempurl%
\url{https://arxiv.org/abs/2404.04925}
\showURL{%
\tempurl}


\bibitem[Sapir(1929)]%
        {Sapir}
\bibfield{author}{\bibinfo{person}{E. Sapir}.} \bibinfo{year}{1929}\natexlab{}.
\newblock \showarticletitle{The Status of Linguistics as a Science}.
\newblock \bibinfo{journal}{\emph{Language}} \bibinfo{volume}{5}, \bibinfo{number}{4} (\bibinfo{year}{1929}), \bibinfo{pages}{207--214}.
\newblock
\showISSN{00978507, 15350665}
\urldef\tempurl%
\url{http://www.jstor.org/stable/409588}
\showURL{%
\tempurl}


\bibitem[Shi et~al\mbox{.}(2023)]%
        {MGSM}
\bibfield{author}{\bibinfo{person}{Freda Shi}, \bibinfo{person}{Mirac Suzgun}, \bibinfo{person}{Markus Freitag}, \bibinfo{person}{Xuezhi Wang}, \bibinfo{person}{Suraj Srivats}, \bibinfo{person}{Soroush Vosoughi}, \bibinfo{person}{Hyung~Won Chung}, \bibinfo{person}{Yi Tay}, \bibinfo{person}{Sebastian Ruder}, \bibinfo{person}{Denny Zhou}, \bibinfo{person}{Dipanjan Das}, {and} \bibinfo{person}{Jason Wei}.} \bibinfo{year}{2023}\natexlab{}.
\newblock \showarticletitle{Language models are multilingual chain-of-thought reasoners}. In \bibinfo{booktitle}{\emph{The Eleventh International Conference on Learning Representations}}. \bibinfo{publisher}{International Conference on Learning Representations}, \bibinfo{address}{Kigali, Rwanda}, \bibinfo{numpages}{20}~pages.
\newblock
\urldef\tempurl%
\url{https://openreview.net/forum?id=fR3wGCk-IXp}
\showURL{%
\tempurl}


\bibitem[Shi et~al\mbox{.}(2024)]%
        {Detecting_Pretraining}
\bibfield{author}{\bibinfo{person}{Weijia Shi}, \bibinfo{person}{Anirudh Ajith}, \bibinfo{person}{Mengzhou Xia}, \bibinfo{person}{Yangsibo Huang}, \bibinfo{person}{Daogao Liu}, \bibinfo{person}{Terra Blevins}, \bibinfo{person}{Danqi Chen}, {and} \bibinfo{person}{Luke Zettlemoyer}.} \bibinfo{year}{2024}\natexlab{}.
\newblock \showarticletitle{Detecting Pretraining Data from Large Language Models}. In \bibinfo{booktitle}{\emph{The Twelfth International Conference on Learning Representations}}. \bibinfo{publisher}{International Conference on Learning Representations}, \bibinfo{address}{Vienna, Austria}, \bibinfo{numpages}{18}~pages.
\newblock
\urldef\tempurl%
\url{https://openreview.net/forum?id=zWqr3MQuNs}
\showURL{%
\tempurl}


\bibitem[Vygotsky(1962)]%
        {Vygotsky}
\bibfield{author}{\bibinfo{person}{Lev~S. Vygotsky}.} \bibinfo{year}{1962}\natexlab{}.
\newblock \bibinfo{booktitle}{\emph{Thought and Language}}.
\newblock \bibinfo{publisher}{MIT Press}, \bibinfo{address}{Cambridge, MA}.
\newblock


\bibitem[Wang et~al\mbox{.}(2024a)]%
        {Sseaeval}
\bibfield{author}{\bibinfo{person}{Bin Wang}, \bibinfo{person}{Zhengyuan Liu}, \bibinfo{person}{Xin Huang}, \bibinfo{person}{Fangkai Jiao}, \bibinfo{person}{Yang Ding}, \bibinfo{person}{AiTi Aw}, {and} \bibinfo{person}{Nancy Chen}.} \bibinfo{year}{2024}\natexlab{a}.
\newblock \showarticletitle{{S}ea{E}val for Multilingual Foundation Models: From Cross-Lingual Alignment to Cultural Reasoning}. In \bibinfo{booktitle}{\emph{Proceedings of the 2024 Conference of the North American Chapter of the Association for Computational Linguistics: Human Language Technologies (Volume 1: Long Papers)}}, \bibfield{editor}{\bibinfo{person}{Kevin Duh}, \bibinfo{person}{Helena Gomez}, {and} \bibinfo{person}{Steven Bethard}} (Eds.). \bibinfo{publisher}{Association for Computational Linguistics}, \bibinfo{address}{Mexico City, Mexico}, \bibinfo{pages}{370--390}.
\newblock
\href{https://doi.org/10.18653/v1/2024.naacl-long.22}{doi:\nolinkurl{10.18653/v1/2024.naacl-long.22}}


\bibitem[Wang et~al\mbox{.}(2024c)]%
        {probing-emergence-cross-lingual}
\bibfield{author}{\bibinfo{person}{Hetong Wang}, \bibinfo{person}{Pasquale Minervini}, {and} \bibinfo{person}{Edoardo Ponti}.} \bibinfo{year}{2024}\natexlab{c}.
\newblock \showarticletitle{Probing the Emergence of Cross-lingual Alignment during {LLM} Training}. In \bibinfo{booktitle}{\emph{Findings of the Association for Computational Linguistics: ACL 2024}}, \bibfield{editor}{\bibinfo{person}{Lun-Wei Ku}, \bibinfo{person}{Andre Martins}, {and} \bibinfo{person}{Vivek Srikumar}} (Eds.). \bibinfo{publisher}{Association for Computational Linguistics}, \bibinfo{address}{Bangkok, Thailand}, \bibinfo{pages}{12159--12173}.
\newblock
\href{https://doi.org/10.18653/v1/2024.findings-acl.724}{doi:\nolinkurl{10.18653/v1/2024.findings-acl.724}}


\bibitem[Wang et~al\mbox{.}(2024d)]%
        {wang2024incharacter}
\bibfield{author}{\bibinfo{person}{Xintao Wang}, \bibinfo{person}{Yunze Xiao}, \bibinfo{person}{Jen tse Huang}, \bibinfo{person}{Siyu Yuan}, \bibinfo{person}{Rui Xu}, \bibinfo{person}{Haoran Guo}, \bibinfo{person}{Quan Tu}, \bibinfo{person}{Yaying Fei}, \bibinfo{person}{Ziang Leng}, \bibinfo{person}{Wei Wang}, \bibinfo{person}{Jiangjie Chen}, \bibinfo{person}{Cheng Li}, {and} \bibinfo{person}{Yanghua Xiao}.} \bibinfo{year}{2024}\natexlab{d}.
\newblock \bibinfo{title}{InCharacter: Evaluating Personality Fidelity in Role-Playing Agents through Psychological Interviews}.
\newblock
\showeprint[arxiv]{2310.17976}~[cs.CL]


\bibitem[Wang et~al\mbox{.}(2024b)]%
        {mmlu_pro}
\bibfield{author}{\bibinfo{person}{Yubo Wang}, \bibinfo{person}{Xueguang Ma}, \bibinfo{person}{Ge Zhang}, \bibinfo{person}{Yuansheng Ni}, \bibinfo{person}{Abhranil Chandra}, \bibinfo{person}{Shiguang Guo}, \bibinfo{person}{Weiming Ren}, \bibinfo{person}{Aaran Arulraj}, \bibinfo{person}{Xuan He}, \bibinfo{person}{Ziyan Jiang}, \bibinfo{person}{Tianle Li}, \bibinfo{person}{Max Ku}, \bibinfo{person}{Kai Wang}, \bibinfo{person}{Alex Zhuang}, \bibinfo{person}{Rongqi Fan}, \bibinfo{person}{Xiang Yue}, {and} \bibinfo{person}{Wenhu Chen}.} \bibinfo{year}{2024}\natexlab{b}.
\newblock \bibinfo{title}{MMLU-Pro: A More Robust and Challenging Multi-Task Language Understanding Benchmark}.
\newblock
\showeprint[arxiv]{2406.01574}~[cs.CL]
\urldef\tempurl%
\url{https://arxiv.org/abs/2406.01574}
\showURL{%
\tempurl}


\bibitem[Wei et~al\mbox{.}(2023)]%
        {CoT}
\bibfield{author}{\bibinfo{person}{Jason Wei}, \bibinfo{person}{Xuezhi Wang}, \bibinfo{person}{Dale Schuurmans}, \bibinfo{person}{Maarten Bosma}, \bibinfo{person}{Brian Ichter}, \bibinfo{person}{Fei Xia}, \bibinfo{person}{Ed Chi}, \bibinfo{person}{Quoc Le}, {and} \bibinfo{person}{Denny Zhou}.} \bibinfo{year}{2023}\natexlab{}.
\newblock \bibinfo{title}{Chain-of-Thought Prompting Elicits Reasoning in Large Language Models}.
\newblock
\showeprint[arxiv]{2201.11903}~[cs.CL]
\urldef\tempurl%
\url{https://arxiv.org/abs/2201.11903}
\showURL{%
\tempurl}


\bibitem[Whorf(2012)]%
        {Whorf}
\bibfield{author}{\bibinfo{person}{Benjamin~Lee Whorf}.} \bibinfo{year}{2012}\natexlab{}.
\newblock \bibinfo{booktitle}{\emph{Language, Thought, and Reality: Selected Writings of Benjamin Lee Whorf}}.
\newblock \bibinfo{publisher}{The MIT Press}, \bibinfo{address}{Cambridge, MA}.
\newblock
\showISBNx{9780262517751}
\urldef\tempurl%
\url{http://www.jstor.org/stable/j.ctt5hhbx2}
\showURL{%
\tempurl}


\bibitem[Yu et~al\mbox{.}(2024)]%
        {reasoning_survey}
\bibfield{author}{\bibinfo{person}{Fei Yu}, \bibinfo{person}{Hongbo Zhang}, \bibinfo{person}{Prayag Tiwari}, {and} \bibinfo{person}{Benyou Wang}.} \bibinfo{year}{2024}\natexlab{}.
\newblock \showarticletitle{Natural Language Reasoning, A Survey}.
\newblock \bibinfo{journal}{\emph{ACM Comput. Surv.}} \bibinfo{volume}{56}, \bibinfo{number}{12}, Article \bibinfo{articleno}{304} (\bibinfo{date}{Oct.} \bibinfo{year}{2024}), \bibinfo{numpages}{39}~pages.
\newblock
\showISSN{0360-0300}
\href{https://doi.org/10.1145/3664194}{doi:\nolinkurl{10.1145/3664194}}


\bibitem[Zeng et~al\mbox{.}(2024)]%
        {multilingual_perplexity}
\bibfield{author}{\bibinfo{person}{Hongchuan Zeng}, \bibinfo{person}{Hongshen Xu}, \bibinfo{person}{Lu Chen}, {and} \bibinfo{person}{Kai Yu}.} \bibinfo{year}{2024}\natexlab{}.
\newblock \showarticletitle{Multilingual Brain Surgeon: Large Language Models Can Be Compressed Leaving No Language behind}. In \bibinfo{booktitle}{\emph{Proceedings of the 2024 Joint International Conference on Computational Linguistics, Language Resources and Evaluation (LREC-COLING 2024)}}, \bibfield{editor}{\bibinfo{person}{Nicoletta Calzolari}, \bibinfo{person}{Min-Yen Kan}, \bibinfo{person}{Veronique Hoste}, \bibinfo{person}{Alessandro Lenci}, \bibinfo{person}{Sakriani Sakti}, {and} \bibinfo{person}{Nianwen Xue}} (Eds.). \bibinfo{publisher}{ELRA and ICCL}, \bibinfo{address}{Torino, Italia}, \bibinfo{pages}{11794--11812}.
\newblock
\urldef\tempurl%
\url{https://aclanthology.org/2024.lrec-main.1030}
\showURL{%
\tempurl}


\bibitem[Zhang et~al\mbox{.}(2023)]%
        {bilingual}
\bibfield{author}{\bibinfo{person}{Xiang Zhang}, \bibinfo{person}{Senyu Li}, \bibinfo{person}{Bradley Hauer}, \bibinfo{person}{Ning Shi}, {and} \bibinfo{person}{Grzegorz Kondrak}.} \bibinfo{year}{2023}\natexlab{}.
\newblock \showarticletitle{Don`t Trust {C}hat{GPT} when your Question is not in {E}nglish: A Study of Multilingual Abilities and Types of {LLM}s}. In \bibinfo{booktitle}{\emph{Proceedings of the 2023 Conference on Empirical Methods in Natural Language Processing}}, \bibfield{editor}{\bibinfo{person}{Houda Bouamor}, \bibinfo{person}{Juan Pino}, {and} \bibinfo{person}{Kalika Bali}} (Eds.). \bibinfo{publisher}{Association for Computational Linguistics}, \bibinfo{address}{Singapore}, \bibinfo{pages}{7915--7927}.
\newblock
\href{https://doi.org/10.18653/v1/2023.emnlp-main.491}{doi:\nolinkurl{10.18653/v1/2023.emnlp-main.491}}


\bibitem[Zhao et~al\mbox{.}(2024a)]%
        {Tracing_the_roots}
\bibfield{author}{\bibinfo{person}{Xin Zhao}, \bibinfo{person}{Naoki Yoshinaga}, {and} \bibinfo{person}{Daisuke Oba}.} \bibinfo{year}{2024}\natexlab{a}.
\newblock \showarticletitle{Tracing the Roots of Facts in Multilingual Language Models: Independent, Shared, and Transferred Knowledge}. In \bibinfo{booktitle}{\emph{Proceedings of the 18th Conference of the European Chapter of the Association for Computational Linguistics (Volume 1: Long Papers)}}, \bibfield{editor}{\bibinfo{person}{Yvette Graham} {and} \bibinfo{person}{Matthew Purver}} (Eds.). \bibinfo{publisher}{Association for Computational Linguistics}, \bibinfo{address}{St. Julian{'}s, Malta}, \bibinfo{pages}{2088--2102}.
\newblock
\urldef\tempurl%
\url{https://aclanthology.org/2024.eacl-long.127}
\showURL{%
\tempurl}


\bibitem[Zhao et~al\mbox{.}(2024b)]%
        {Multilingualism}
\bibfield{author}{\bibinfo{person}{Yiran Zhao}, \bibinfo{person}{Wenxuan Zhang}, \bibinfo{person}{Guizhen Chen}, \bibinfo{person}{Kenji Kawaguchi}, {and} \bibinfo{person}{Lidong Bing}.} \bibinfo{year}{2024}\natexlab{b}.
\newblock \showarticletitle{How do Large Language Models Handle Multilingualism?}. In \bibinfo{booktitle}{\emph{The Thirty-eighth Annual Conference on Neural Information Processing Systems}}. \bibinfo{publisher}{Neural Information Processing Systems}, \bibinfo{address}{Vancouver, Canada}, \bibinfo{numpages}{24}~pages.
\newblock
\urldef\tempurl%
\url{https://openreview.net/forum?id=ctXYOoAgRy}
\showURL{%
\tempurl}


\end{thebibliography}
